\documentclass[journal]{IEEEtran}
\usepackage[ruled]{algorithm2e}
\usepackage{epsfig} 
\usepackage{mathrsfs}
\usepackage{verbatim}
\usepackage{graphicx}
\usepackage{float}
\usepackage{graphics} 
\usepackage{epsfig} 
\usepackage{times} 
\usepackage{amsmath} 
\usepackage{amssymb}  
\usepackage{bm}
\usepackage{cite}
\usepackage{graphicx}
\usepackage[numbers,sort&compress]{natbib}
\usepackage{booktabs}
\usepackage{color}
\usepackage{indentfirst}
\usepackage{soul}
\usepackage{multirow}
\usepackage{colortbl}  
\usepackage{xcolor}
\usepackage{framed}
\usepackage{lipsum}
\usepackage{balance}
\usepackage{enumerate}
\usepackage{cleveref}
\usepackage{booktabs}
\usepackage{threeparttable}
\colorlet{shadecolor}{yellow!20}
\newtheorem{Theorem}{Theorem}
\newtheorem{Remark}{Remark}
\newtheorem{Assumption}{Assumption}

\ifCLASSINFOpdf
\else
\fi
\begin{document}
%
\title{Towards Practical Autonomous Flight Simulation for Flapping Wing Biomimetic Robots with Experimental Validation}
%
%
%

\author{Chen~Qian,
        Yongchun~Fang,~\IEEEmembership{Senior Member,~IEEE},
        Fan~jia,
        Jifu~Yan,
        Yiming~Liang 
        and Tiefeng~Li
\thanks{This work was supported by National Natural Science Foundation
of China under Grant 62233011, T2125009, 92048302. (Corresponding authors: Yongchun
Fang and Tiefeng~Li.)

C. Qian and Y. Liang are with Interdisciplinary Innovation Research Centers, Intelligent Robotic Research Center, Zhejiang Laboratory, Hangzhou
311100, China (e-mail: qianc@zhejianglab.com, liangym@zhejianglab.com).

Y. Fang, F. Jia, J. Yan, are with College of Artificial Intelligence, Nankai University, and Institute of Robotics and Automatic Information
Systems, Nankai University, Tianjin, 300353, China (e-mail: fangyc@nankai.edu.cn, fanjia@mail.nankai.edu.cn).

T. Li is with School of Aeronautics and Astronautics, ZheJiang University, Hangzhou 310012, China (e-mail: litiefeng@zju.edu.cn).

}
}

%
%

\markboth{Preparing submission, Dec 15th 2022}%
{Shell \MakeLowercase{\textit{et al.}}: Bare Demo of IEEEtran.cls for IEEE Journals}
%



\maketitle

\begin{abstract}
Tried-and-true flapping wing robot simulation is essential in developing flapping wing mechanisms and algorithms.
This paper presents a novel application-oriented flapping wing platform,  highly compatible with various mechanical designs and adaptable to different robotic tasks.
First, the blade element theory and the quasi-steady model are put forward to compute the flapping wing aerodynamics based on wing kinematics.
Translational lift, translational drag, rotational lift, and added mass force are all considered in the computation.
Then we use the proposed simulation platform to investigate the 
passive wing rotation and the wing-tail interaction phenomena of a particular flapping-wing robot.
With the help of the simulation tool and a novel statistic based on dynamic
differences from the averaged system, 
several behaviors display their essence by investigating the flapping wing robot dynamic characteristics.
After that, the attitude tracking control problem and the positional trajectory tracking problem are both
overcome by robust control techniques. Further comparison simulations reveal that the proposed control algorithms compared with other existing 
ones show apparent superiority.
What is more, with the same control algorithm and parameters tuned in simulation, we conduct real flight experiments on a
self-made flapping wing robot, and obtain similar results from the proposed simulation platform.
In contrast to existing simulation tools, the proposed one is compatible with most existing flapping wing robots,
and can inherently drill into each subtle behavior in corresponding applications by observing aerodynamic forces and torques on each blade element.
\end{abstract}


%
\IEEEpeerreviewmaketitle

\section{Introduction}
\IEEEPARstart{S}{imulating} flapping wing flight within practical robotic tasks 
can help analyze and build flapping wing robots, as well as various associated algorithms.
On the other hand, possibly only through application-oriented environmental interactions, 
sufficiently precise modelings, 
platforms allowing nuanced observations, 
we can decipher flapping flight to a novel stage. 
Furthermore, if a simulation platform can stimulate the 
development of different flapping wing robots,
it should be sufficiently compatible and expandable to different robot designs.

Compared with state-of-art conventional unmanned aerial vehicle studies, 
flapping wing robot researches remain in a relatively elementary stage.
Researchers aspire to bridge the gap between aerodynamics studies, 
mechanism studies, and robotic studies \cite{Tijmons-2017, Kar-2018, Tu-2020, Rodriguez-2021, Zufferey-2021, Chen-2022, Wu-2022}.
In \cite{Kar-2018}, Kar\'{a}sek \emph{et al.} develop an X-shape wings tailless robot with servos rotating the flapping mechanism
and asynchronous bilateral flapping wing actuation,
which can imitate the rapid escape maneuvers of flies.
Tu \emph{et al.} design a hummingbird-inspired, at-scale, tail-less flapping wing micro aerial vehicle, which
independently controls its wings with a total of only two directly driven motors \cite{Tu-2020}.
Zufferey \emph{et al.} develop an Eagle-inspired Flapping-wing robot E-Flap that can carry a 100\% of the payload,
which has two aero-elastic wings \cite{Zufferey-2021}. 
Chen \emph{et al.} design a novel bat-style flapping wing robot, which
couples or decouples the morphing and flapping, and can generate a bilateral asymmetric downstroke affording high rolling agility \cite{Chen-2022}.
These prototypes possess considerable disparities. Flapping wing robots and algorithms developments strongly depend on experimental studies
and empirical knowledge concluded from data-rich conventional planes or drones, meanwhile,
flapping wing aerodynamic investigations and specific robotic task studies are virtually disengaged. 

Flapping wing flight simulation can provide profound inspiration and instructions for real flight tasks.
In \cite{Fei-2019}, Fei \emph{et al.} provide an open-source high fidelity dynamic simulation for 
their flapping wing robot. After applying system identification,  the same
flight performance can be achieved on the robot by directly
implementing the controller in simulation. 
However, the aerodynamics computation of wings is highly simplified, 
which hinders the observation of each blade element aerodynamics in simulation, 
and unable to simulate subtle aerodynamic behaviors such as center of pressure (CoP) 2D positional changing on the wing surface.
Furthermore, the tailless design of their robot makes the attendant simulation rarely consider wing-tail or wing-wing aerodynamic interaction.
This hinders its application for many other flapping wing robots with tails or tandem wings.
Although their simulation can be integrated into the Robot Operating System (ROS) and Gazebo, 
many applications involving environment building, perception, and interaction tasks remain elusive and obviously require cumbersome additional programming. 
Thus, substantial improvement of this simulation platform is indispensable and imperative.
In \cite{Orlowski-2021}, Orlowski \emph{et al.} use a system of three rigid bodies including the body and two wings to 
simulate the flapping wing robot, and further conclude that mass effects
of the wings can exert non-negligible influence on dynamics, stability, and control analyses.
However, there still exists no wing-tail or wing-wing interaction. 
Moreover, the absence of an interface limits its easy-to-use potential for robotic applications.
In \cite{Lopez-2020}, Lopez-Lopez \emph{et al.} propose a simple
but effective analytical model for a specific flapping wings UAV in
longitudinal gliding flight, and the corresponding environment has been built on
Unreal Engine 4 for rendering. The simulation switches between aerodynamic model and 
collision model for their different dynamics, and supports both flapping and gliding flights.
Regretfully, the wing aerodynamics is also possibly over-simplified, such that 
delicate observation is not optional. And the bespoke mode makes the simulation
limited in extensibility.
In \cite{{He-2021}}, He \emph{et al.} establish a simulation model for flapping wing robot
longitudinal motion. Nevertheless, the simulation is dedicated to the control task, 
and due to the longitudinal limitation, it is difficult to implement for most flapping wing robot tasks.

In conclusion, the practical flapping wing flight simulation platform should have the following characteristics:
\begin{enumerate}
  \item The simulation platform should consider both the distinctive modeling aerodynamics and the 
  multi-body dynamics. 
  The wing-wing and wing-tail aerodynamics interactions should also be investigated.
  \item The platform can be compatible with different modeling configurations. 
  And different designs can be losslessly transformed into its simulation counterpart.
  \item In the simulation, different robotic tasks can be performed. 
  Furthermore, the platform ideally provides a user-friendly interface for widely used robotic software or algorithms that can be smoothly integrated into online programs for real flight.
\end{enumerate}

The main purpose of this paper is to realize the simulation platform and validate it with experiments.
Quasi-steady aerodynamics model and blade element method are the common recipes for modeling flapping-wings.  
Quasi-steady models are usually used to describe the unsteady process with basic principles and empirical formulae,
which provide a tractable means of calculating instantaneous forces from defined or generated wing kinematics \cite{Sane-2002}.
When using blade element theory, aerodynamic forces computations are performed on spanwisely divided wing strips, 
which naturally considers wing aerodynamics spatial heterogeneity.
Due to the computation simplicity, relatively high fidelity and high compatibility to control dynamic models, 
many works involving flapping wing aerodynamics implement these methods \cite{Ansari-2006, Deng-2006, Chukewad-2021, Chen-2021}.
Most simulation platforms are first tested with control tasks, meanwhile, the control problem 
is also the main distinctive part of flapping wing robotic tasks \cite{Fei-2019, Orlowski-2021, Lopez-2020, He-2021, Rifai, HeWI, Paranjape}. Thus, we use attitude tracking and trajectory tracking tasks to test the proposed simulation platform, 
meanwhile, other conventional robotic tasks such as perception and navigation can also be straightforwardly performed.  

This work develops an open-source simulation platform satisfying the aforementioned three characteristics.
The contributions can be concluded into the following four aspects:
\begin{enumerate}
  \item Most existing flapping wing robots can be realized on the developed simulation platform
  without any complicated adaptation, 
  which can help design, modify and validate both the robot itself and the algorithm 
  thereon. By bridging the gap between algorithms designed for an individual robots in a bespoke fashion,
  the proposed platform can accelerate the flapping wing design and test process. 

  \item In contrast to existing simulations, aerodynamic force computations in the proposed simulation are losslessly implemented, allowing us to investigate a variety of subtle behaviors in robotic tasks.
  The simulation fidelity is sufficiently high that both algorithms and parameters 
  can be directly applied in real flight even without identification.
  This can help us understand unexpected behaviors owing to complicated flapping wing aerodynamics, which 
  can also be utilized to explore natural flight behaviors.
  
  \item A highly extensible real flight program framework is developed. The real flight algorithm shares the 
  same algorithms and parameters in simulation, which facilitates applying the optimal algorithms and parameters obtained from simulation results. 

  \item A novel and theoretically more effective statistic capturing the flapping wing oscillating dynamics is proposed.

\end{enumerate}

The remainder of this paper is organized as follows. Section \uppercase\expandafter{\romannumeral2} explains the basic aerodynamics used in the simulation. Section \uppercase\expandafter{\romannumeral3} studies the specific flapping wing robot aerodynamics performance, where a self-made X-shape tailed flapping wing robot is used as an example. In Section \uppercase\expandafter{\romannumeral4}, practical robotic applications are performed. We respectively study the attitude tracking and positional trajectory tracking problem for the robot, and compare them with other existing controllers to show their superiority.
Then in Section \uppercase\expandafter{\romannumeral5}, the real counterpart robot is built,
 and the real flight program framework is developed. The real flight experiments are performed to validate the simulation results.
Finally, Section \uppercase\expandafter{\romannumeral6} concludes this work.

\begin{figure}[t]
\centering
\includegraphics[width=3in]{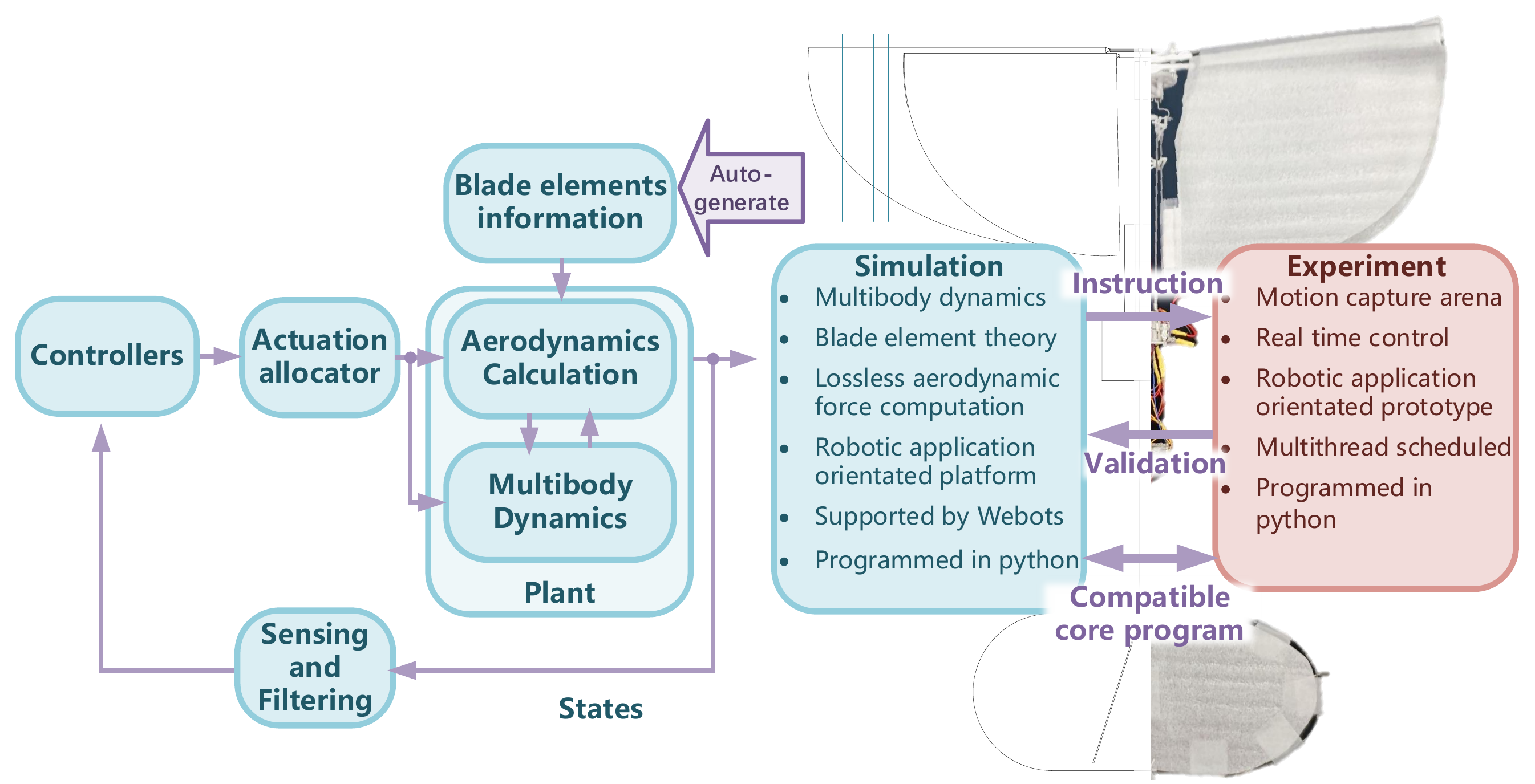}
\caption{Flapping wing robot simulation platform and experiment.}
\label{figure:coredes}
\end{figure}

\section{Flapping Wing Aerodynamics}
\subsection{Quasi-steady Aerodynamics}
The unsteady aerodynamics force mechanisms in flapping flight such as leading-edge vortex, added mass, wake capture, rotational circulation, and clap-and-fling effect should be well considered in order to obtain an accurate aerodynamic or dynamic model.
Conducting a large amount of simulations require an easily computed model.
The combination of the blade element theory and the quasi-steady models can actually strike a satisfactory balance between simplicity and fidelity, which depends on the instantaneous wing kinematics such as velocities and accelerations, as well as the wing morphology. From another perspective, they are modeled by the assumption of inherently time-independent fluid dynamic mechanisms \cite{Lee-2016,Chin-2016,Sane-2003}.
Specifically, the local force acting on a wing strip, or blade element, can be integrated over the wingspan to obtain the aerodynamic forces acting on the wings. The wake capture mechanism is excluded in our simulation for the following two reasons: its inherent
dependence on the airflow history, and its insignificant contribution to improving flying efficiency~\cite{Phan-2019}.
Generally, the instantaneous aerodynamics force can be calculated in the following manner:
\begin{align}
\label{eq:instAeroForce}
\bm F = {\bm F_{{\rm{t,lift}}}} + {\bm F_{{\rm{t,drag}}}} + {\bm F_r} + {\bm F_a}
\end{align}
where ${\bm F_{{\rm{t,lift}}}}$ is the translational lift, acting perpendicular to the wing velocity, ${\bm F_{{\rm{t,drag}}}}$ is the  translational drag, opposing the wing velocity, ${\bm F_r}$ is rotational force caused by the rotational circulation, associated with the wing pronation or supination, and ${\bm F_a}$ is the added mass force, which is associated with the wing acceleration. Based on the quasi-steady model given in \cite{Chin-2016}, the aforementioned forces interact synergistically, but for conciseness can be calculated as
\begin{align}
 &{\rm Mag}\left( {{\bm F_{{\rm{t,lift}}}}} \right) = \sum {\frac{1}{2}\rho c{{\left\| {{\bm u_w}} \right\|}^2}} {C_{Lt}}\left( \alpha  \right)\Delta r, \nonumber\\ 
&{\rm Mag}\left(  {{\bm F_{{\rm{t,drag}}}}} \right) = \sum {\frac{1}{2}\rho c{{\left\| {{\bm u_w}} \right\|}^2}} {C_{Dt}}\left( \alpha  \right)\Delta r, \nonumber\\ 
 &{\rm Mag}\left(  {{\bm F_r}} \right) = \sum {\rho {c^2}\left\| {\dot \alpha } \right\|\left\| {{\bm u_w}} \right\| {C_r}\Delta r}, \nonumber\\ 
 \label{eq:instadded}
 &{\rm Mag}\left( {{\bm F_a}}\right)  = \sum {\frac{{\rho \pi {c^2}}}{4}} \left\{ {\frac{{{\bm u_w} \cdot {{\dot {\bm u}}_w}}}{{\left\| {{\bm u_w}} \right\|}}\sin \alpha  + \left\| {{{\dot {\bm u}}_w}} \right\|\alpha \cos \alpha } \right\}\Delta r
\end{align}
where ${\rm Mag}(\star)$ indicates the magnitude of the corresponding force, which can be positive or negative depended on the angle of attack (AoA) $\alpha$, $\rho$ is the air density, $c$ is the chord length, $\bm u_w$ is the wing-strip (\emph{i.e.} the blade element) effective velocity, $\Delta r$ is the wing-strip width, $C_{Lt}(\alpha)$ and $C_{Dt}(\alpha)$  are the translational lift and drag coefficients, respectively, $C_{r}$ is the rotational coefficient, and $\alpha$ is the effective AoA, which is calculated by incorporating with the resultant rotation from the torso to the wing chord. Each blade element may have a different AoA due to the temporally and spatially varying flow, such that the local orientations of lift and drag also differ.

Based on the experiments reported in \cite{Pre1}, besides the free-stream velocity, the flapping wing induced velocity also obviously affects the tail aerodynamic forces. 
Since the flapping wing induced wake is complex and fast with respect to the tail movement,  we can use the actuator disk model to estimate the average flapping wing induced velocity instead. This average velocity ${\bm u_i}$ is considered in the opposite direction to the resultant force of the translational lifts generated by wings:
\begin{align}
\label{eq:enduced}
{\bm u_i} = \frac{1}{2}\frac{{\Sigma {{\bm F}_{\rm{t,lift}}}}}{{\left\| {\Sigma {{\bm F}_{\rm{t,lift}}}} \right\|}}\sqrt {\frac{{\left\| {\Sigma {{\bm F}_{\rm{t,lift}}}} \right\|}}{{{S_d}}}\frac{1}{{2\rho }}}
\end{align}
where ${\Sigma {{\bm F}_{\rm{t,lift}}}}$ is the resultant force of translational lifts, and $S_d$ is the actuator disk area.


\subsection{Blade Elements}
The concept of blade element is to separate the wing spanwisely into several wing strips. 
In order to determine the chord length of each strip, we provide a method to automatically extract them from 3D-model files\footnote{See source codes at https://github.com/Chainplain/BladeEleBuilder.}.
We first get the vertices points cloud of the wing,
then use a linear interpolation algorithm to obtain a functional description of the wing edge.
Subsequently, we use the polynomial function to help generate an appropriate quantity of aequilate blade elements.  

First, we demonstrate our way to formulate the wing strip effective incident flow velocity $\bm u_w$. Although there are several other impact factors such as the induced downwash flow \cite{Sane-2003}, the jet induced by the shed vortexes \cite{Wu-2005}, \emph{etc.}, in order to make the model more easily computed, we follow the assumption that any span-wise component of the
relative velocity has no effect on the wing forces \cite{Ellington-1999}, and neglect the downwash flow and the vortex shedding, hence we can concentrate on the body motion induced velocity (both translational $\bm v_{Bt}$ and rotational $\bm v_{Br}$), the free flow $\bm U_{\infty}$, and the wing motion induced velocity $\bm v_{W}$. The effective velocity $\bm u_w$ is then given by

\begin{align}
\label{eq:effvel}
{\bm u_\omega } = & {\bm u_{\omega r}} - \hat {\bm n}_S\left( {\hat {\bm n}_S \cdot {\bm u_{\omega r}}} \right),\\
{\bm u_{\omega r}} \triangleq &  {\bm U_\infty } - {\bm v_W} - {\bm v_{Bt}} - {\bm v_{Br}} \nonumber
\end{align}
where $\hat {\bm n}_S$ is the normal vector of the plane that contains the chord, and is also perpendicular to the wing rotation axis.

\begin{figure}[t]
\centering
\includegraphics[width=2.7in]{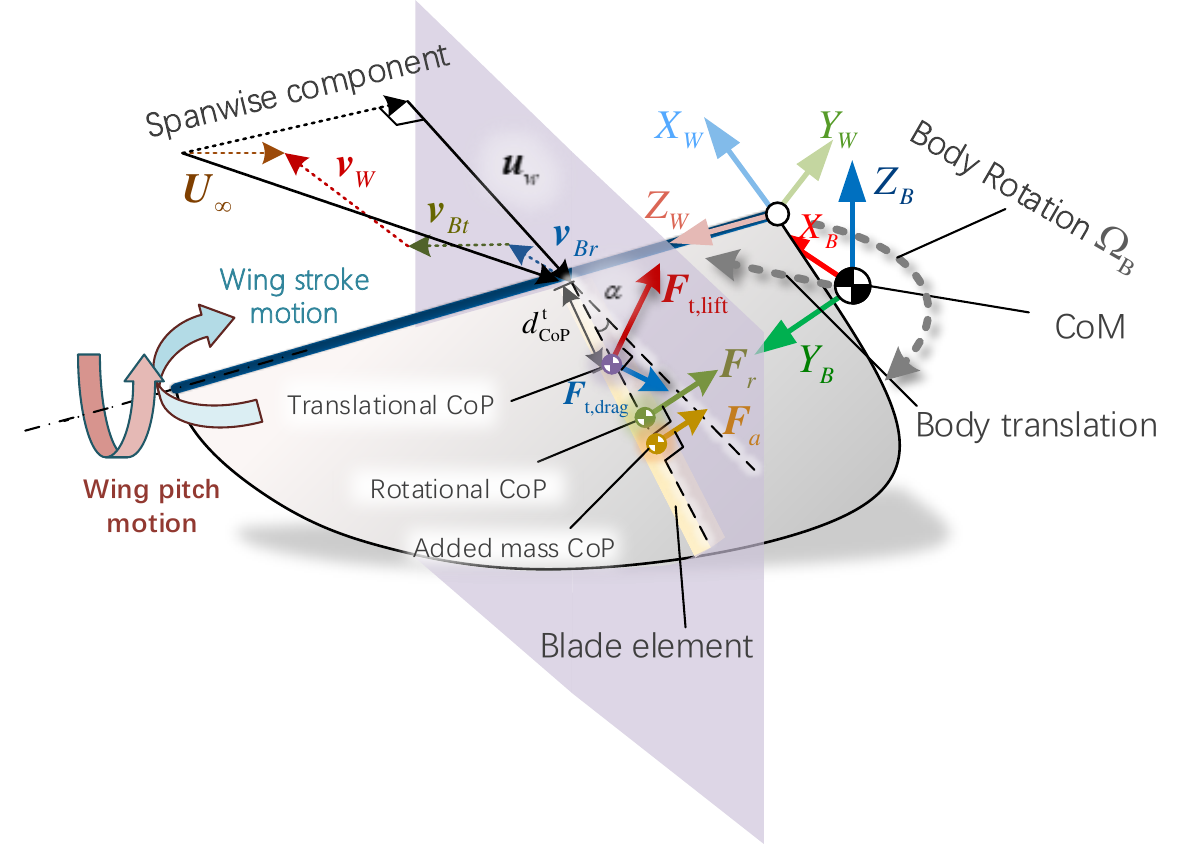}
\caption{The schematic of the flapping wing aerodynamics, where CoP means the center of pressure, and CoM means the center of mass.}
\label{figure:plane}
\end{figure}
The relationship between these velocities is shown in Fig.\ref{figure:plane}.
And the effective AoA $\alpha$ is the angle between the chord and the effective incident velocity, which is shown as 
\begin{align}
\label{eq:alphaCal}
\alpha  = \arccos \frac{{{\bm u_w} \cdot \hat {\bm c}}}{{\left\| {{\bm u_w}} \right\|\left\| {\hat {\bm c}} \right\|}}
\end{align}
where $\hat {\bm c}$ is the vector along the chord direction. Note that all the velocities are resolved at the leading edge even if the force act on the wing strip CoP, to accentuate that flapping translation refers to an airfoil revolving around a central axis \cite{Sane-2003}. And based on \cite{Lee-2016}, we can specify the relationship between the coefficients and the AoA: 
\begin{align}
\label{eq:AeroCoef}
 C_{Lt} & =A_L\sin(2\alpha),\\
C_{Dt} & =C_{D_0}+A_D[1-\cos(2\alpha)]
\end{align}
where the coefficients $C_{Lt}$ and $C_{Dt}$ are both functions of Reynolds number $Re$:
\begin{align}
A_L &= 1.966 -3.94Re^{-0.429}, \nonumber\\
A_D &=1.873 - 3.14Re^{-0.369}, \nonumber\\
C_{D_0} &=0.031+10.48Re^{-0.764}.\nonumber 
\end{align}
As suggested in \cite{Han-2015} and also reported in \cite{Sane-2002}, by using the theoretical value of the standard Kutta–Joukowski theory,
the rotation coefficient can be given as  
\begin{equation}
\label{eq:AeroRotCoef}
{C_r} = \pi \left( {0.75 - {{\hat x}_0}} \right)
\end{equation}
where ${{\hat x}_0}$ is the non-dimensional rotational axis chordwise position, the value of which ranges from 0 to 1.

Second, we consider the aerodynamic moment, which is extremely important for wing passive rotation, and highly sensitive
to changes in the location of the CoP. The moment can be shown similarly to the semi-empirical formulae given in \eqref{eq:instadded}.
However, to pursue consistency and conciseness, we use the deduction given in \cite{Han-2015}, which points out that the CoP due to rotational force is located at 1/2 non-dimensional chord position.
And we further use the model proposed in \cite{Wang-2016} to obtain the CoP due to translational force:
\begin{align}
\label{eq:dtrans}
\hat d_{{\rm{CoP}}}^{\rm{t}} = \frac{1}{\pi }\left| \alpha  \right|,~~~0 \le \alpha  \le \frac{\pi }{2}
\end{align}
Furthermore, since the leading edge and the pitch rotational axis of the FWAV coincide with each other in this work, also according to \cite{Wang-2016}, the CoP due to added mass is located at 9/16 non-dimensional chord position.
Since the clap-and-fling mechanism depends on wings contact, the induced aerodynamic moments can be neglected because their effect is canceled by each other. Once CoPs are determined, we can find the specific lengths of force arms, and corresponding aerodynamic moments can then be straightforwardly computed.

To facilitate the simulation in the sense of conciseness, we need the following assumption.
\begin{Assumption}[Torsional spring]\label{as:Assumption2}
Similar to \cite{{Tu-2018}}, we assume that there exist torsional springs at hinges in pitch rotations, which provides a simplification of the wing flutter and flexibility.
\end{Assumption}

With respect to the tail aerodynamics, we follow the same model as the wings. The only difference is, in the tail aerodynamics, the flapping wing induced velocity $\bm u_i$ is considered, such that we use the resultant velocity of free flow velocity $\bm U_{\infty}$ and induced velocity $\bm u_i$ to calculate the wing effective incident velocity, instead of only the free flow velocity.
The tandem flapping wings can also be simulated in a similar way.

\section{Aerodynamics Simulation Performance}
\begin{figure}[t]
\centering
\includegraphics[width=2.7in]{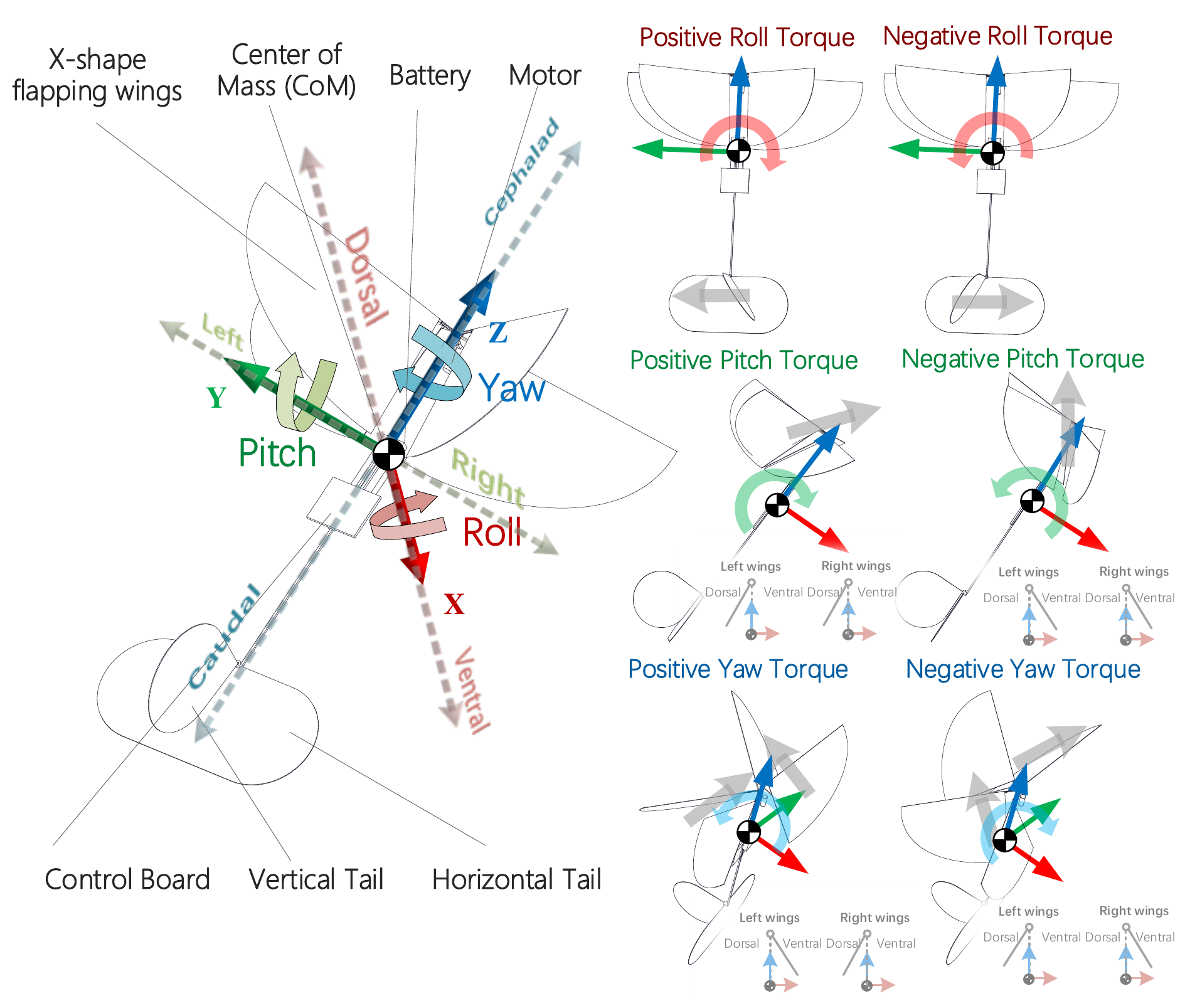}
\caption{Schematic of the flapping wing robot used in this section simulation: three-dimensional body fixed frame consists of three orthogonal axes, where the X-axis is red, Y-axis green, Z-axis blue, and the gray arrow indicates the average lift direction of corresponding wings or tails.}
\label{figure:Basic}
\end{figure}

In order to demonstrate the correctness and accuracy of our simulation, also to manifest its sophistication degree, 
we provide close shots of wing passive rotation and wing-tail interaction phenomena,
which are well-known intractable aerodynamic behaviors in flapping wing simulations.
Furthermore, we also show that the proposed simulation tool can help compute periodically average forces and torques,
which can penetrate the maneuverability characteristics of the robot.
The robot simulated in this section with its body-fixed frame is shown in Fig. \ref{figure:Basic}.
It is an X-wing flapping wing robot, whose equilibrium pitch position is adjustable, with a fixed horizontal tail and a revolute vertical tail. 
The robot can generate roll torques by rotating its vertical tail. 
When left wings and right wings equilibrium pitch position is moved along the same direction, the robot 
can generate pitch torques. 
When left wings and right wings equilibrium pitch position is moved along opposite directions, the robot generates yaw torques.   
Detailed robot parameters are collectively given in TABLE~\ref{tab:PARAS1}.

\subsection{Passive Wing Rotation}
Although the torsional spring assumption, \emph{Assumption \ref{as:Assumption2}}, is relatively simple, due to the detailed aerodynamics simulation, the locomotions and effects are comparatively complex. In most flapping wing robots, powered by motors, and driven by crank-and-rod mechanism, the flapping wing stroke angle are sinusoidal or approximately sinusoidal in time. Thus, we suppose the stroke angle is also sinusoidal in our simulation. In all the simulations, the stroke peak to peak amplitude is set as $\pi / 4~{\rm rad}$ for each wing, the air density is set as $1.29~{\rm kg/m^3}$, the simulation step is $1{~\rm ms}$. 
We also install a stopper at the pitch rotation joint, which can limit the rotation angle within the range of $[-\pi / 4,\pi / 4]~{\rm rad}$, because at the aerodynamic AoA of $\pi / 4~{\rm rad}$, flapping wing is believed to be most efficient. 
In the following, we explain those complex flapping wing behaviors by studying different cases. And according to the characteristics of the flapping wing model used in the simulation, the average Reynolds number $Re$ is 7000. The wing with $14~{\rm cm}$ spanwise length is divided into 40 blade element strips, meanwhile, the vertical and the horizontal tails are both divided into 20 blade element strips, and other number of strips is also optional.
\begin{table}[t]
\caption{Flapping wings simulation parameters}
  \centering
  \label{tab:PARAS1}
\begin{tabular}{@{}llll@{}}
\toprule
Parameters                                                               & Values                         & Parameters                                                             & Values                                              \\ \midrule
mass, $m$                                                                    & 22g                            &wing stop pitch angle                                                       & $\pi /4$                                            \\
\begin{tabular}[c]{@{}l@{}}Reynolds number,\\ $Re$\end{tabular}           & 7000                           & \begin{tabular}[c]{@{}l@{}}pitch balance\\ ~position range\end{tabular} & $\left[ { - \frac{\pi }{4},\frac{\pi }{4}} \right]$ \\
wing span                                                                & 30cm                           & rudder rotation range                                                  & $\left[ { - \frac{\pi }{4},\frac{\pi }{4}} \right]$ \\
air density, $\rho$                                                            & 1.29${\rm{kg}}/{{\rm{m}}^3}$    & ~simulation gap                                                         & 1 ms                                                \\
\begin{tabular}[c]{@{}l@{}}wing torsional\\ ~spring constant\end{tabular} & 0.025${\rm{N}} \cdot {\rm{m}} / {\rm {rad}}$ & \begin{tabular}[c]{@{}l@{}}controller \\ ~~computation gap\end{tabular}  & 10 ms                                               \\ \bottomrule
\end{tabular}
\end{table}
\subsubsection{Low-frequency stroke}
\begin{figure}[t]
\centering
\includegraphics[width=2.7in]{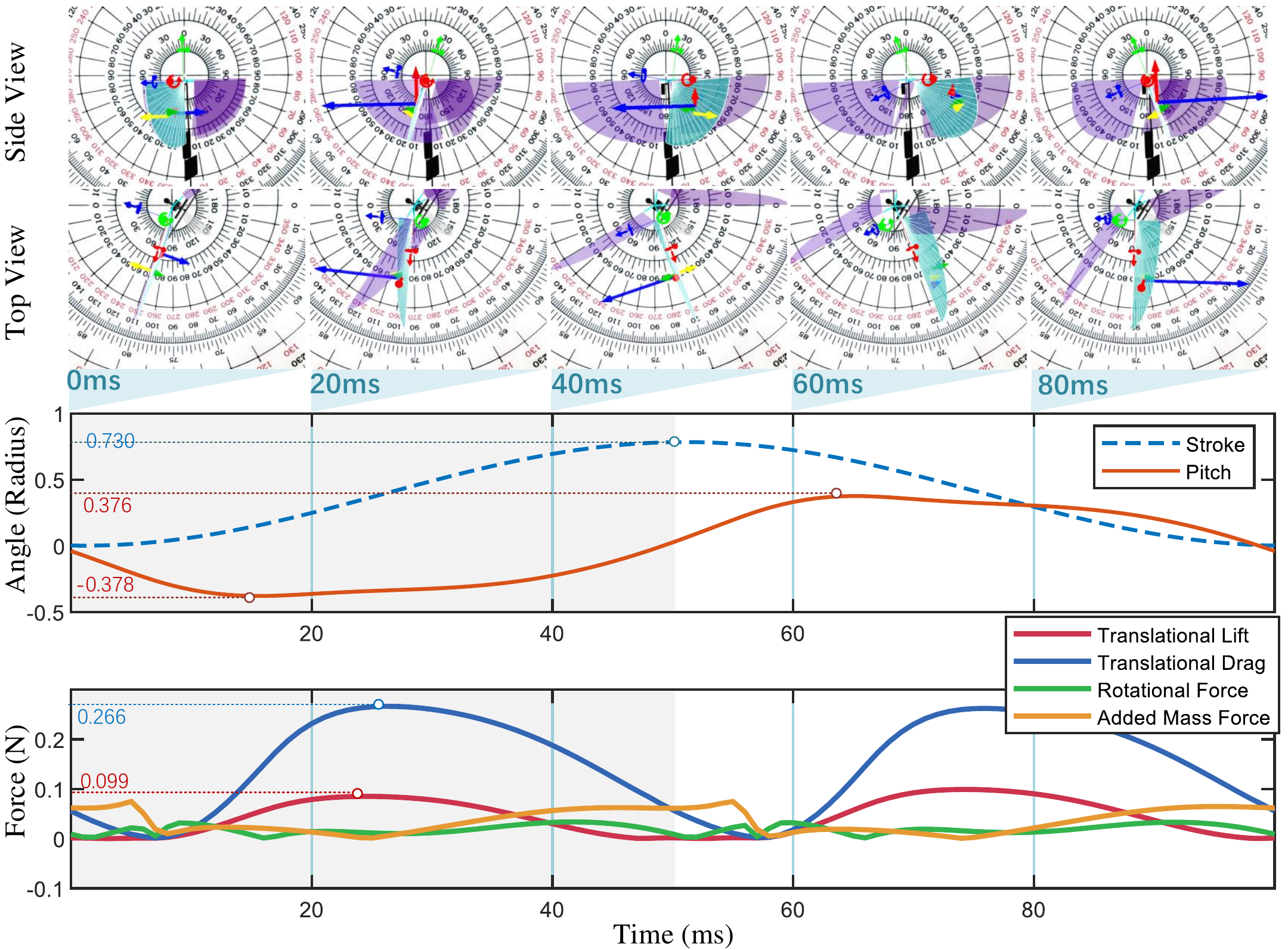}
\caption{Flapping wing simulation results when the torsional spring constant is set as $0.025~{\rm N\cdot m/rad}$, stroke frequency at $10{~\rm Hz}$. 
The curves record the states of the flapping wing marked in light blue. Please note that only the aerodynamics of the upper left wing is displayed to avoid confusion.}
\label{figure:Results1}
\end{figure}

we obtain the simulation result shown in Fig. \ref{figure:Results1}.
The 3D views in the first two rows can provide a direct demonstration of the wing locomotion in a single period of flapping.
Due to the passive rotation mechanism, the pitch angle has a delay of approximately half period. 
Since the stroke frequency is relatively low, the pitch angle is smaller than the most efficient $\pi / 4~{\rm rad}$. 
Because of this, the translation drag amplitude shown in the figure is obviously larger than the one of the translational lift. 
Moreover, we can also see a symmetry between the back-stroke (dark-background part) and the front-stroke (light-background part).
The added mass force becomes obvious, with a maximum magnitude of $0.08~{\rm N}$, when the wing starts to accelerate or decelerate. 
And the rotational force peaks when the wing rotates rapidly. However, there is a rotational force trough instead of a peak when the stroke direction changes, it owes to the drop of the efficient velocity $\bm u_w$. After the transient ascent, the rotational force descends again, as the pitch rotation subsides.
All these behaviors conform to existing works of flapping wing aerodynamics studies,
which manifests the accuracy of our simulation.
\begin{figure}[t]
\centering
\includegraphics[width=2.7in]{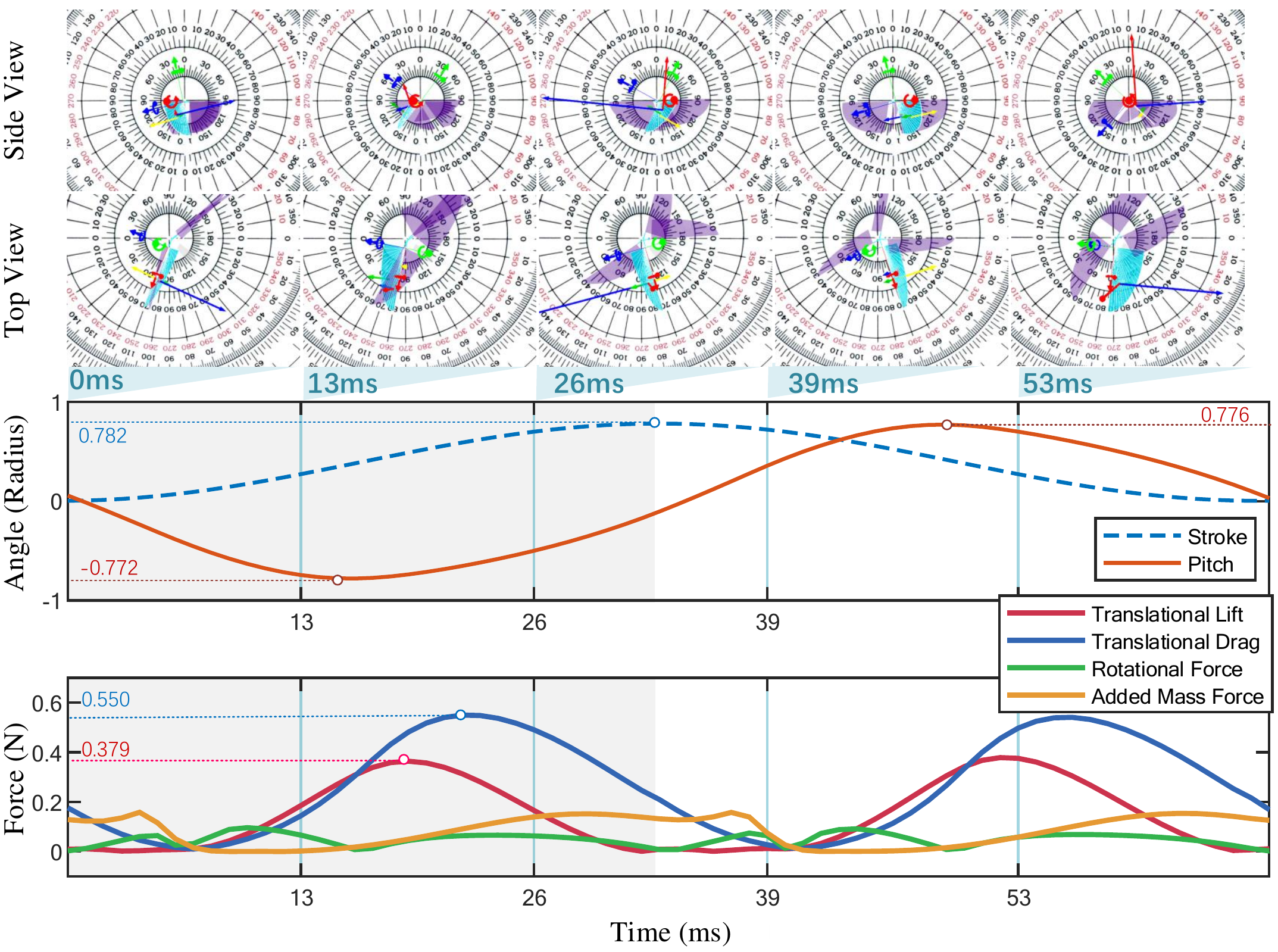}
\caption{Flapping wing simulation results when the torsional spring constant is set as $0.025~{\rm N\cdot m/rad}$, stroke frequency at $15{~\rm Hz}$.
 The curves record states of the flapping wing marked in light blue.}
\label{figure:Results2}
\end{figure}
\begin{figure}[t]
\centering
\includegraphics[width=2.7in]{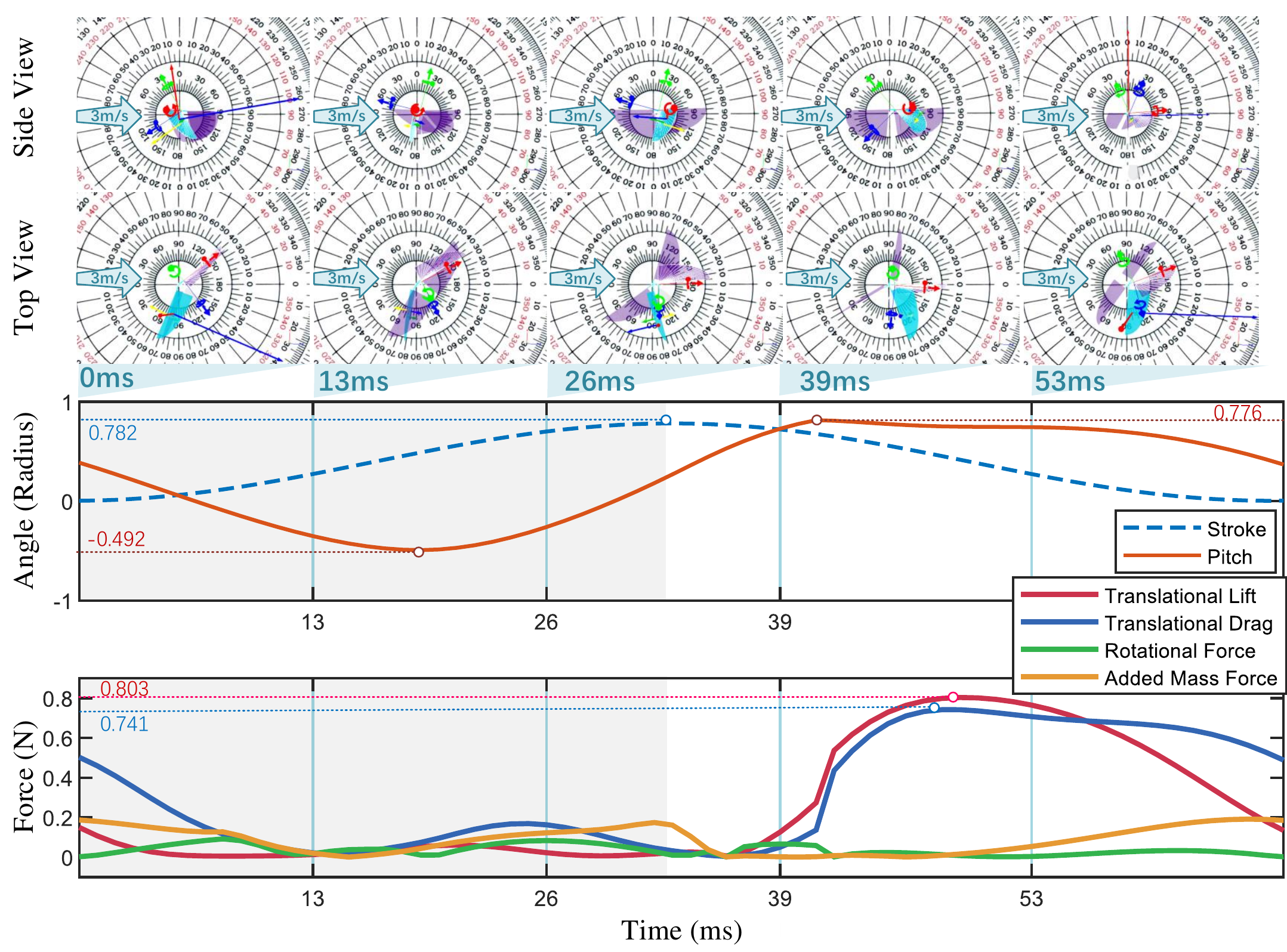}
\caption{ Flapping wing simulation results when the torsional spring constant is set as $0.025~{\rm N\cdot m/rad}$, stroke frequency at $15{~\rm Hz}$, and the flapping wings faces the $3~{\rm m/s}$ wind towards the inertia frame $z$-direction. 
 The curves record states of the flapping wing marked in light blue.}
\label{figure:Results3}
\end{figure}

\subsubsection{High frequency stroke}  When we set the torsional spring constant as $0.025~{\rm N\cdot m/rad}$, and stroke frequency as $15{~\rm Hz}$,
we obtain the simulation result shown in Fig. \ref{figure:Results2}. 
Comparing to the results shown in Fig. \ref{figure:Results1}, the aerodynamic forces are relatively large because of the larger AoA, 
although they demonstrate the similar periodic pattern.  
As the pitch angle reaches the desired $\pi / 4$, the translation lift versus translational drag ratio increases.
Moreover, both the rotational force and the added mass force periodic patterns are kept similar to those in \emph{Case~1}. 
\subsubsection{Side wind} When we set the torsional spring constant as $0.025~{\rm N\cdot m/rad}$, and stroke frequency as $15{~\rm Hz}$,
we obtain the simulation result shown in Fig. \ref{figure:Results3}. Moreover, we additionally exert a wind of $3~{\rm m/s}$ in the inertia frame $z$-direction.
The magnitudes of aerodynamic forces decrease to comparatively small values, when the wing approximately moves along the wind direction. 
Clearly shown in Fig. \ref{figure:Results3}, the maximum pitch angle is 0.776 ${\rm rad}$, while the minimum is -0.492 ${\rm rad}$, 
which manifests the wing passive locomotion asymmetry.
On the other hand, in the front stroke, where the wings move against the wind,
these forces magnitudes are larger than they appear in no wind situations shown in \emph{Case~2}.
Furthermore, the pitch angle keeps the stop angle $\pi /4~{\rm rad}$ for a relatively long time, such that the high lift-drag ratio is also kept for over $25\%$ period.
In conclusion, the side wind causes an asymmetry between front and back stroke. Especially, due to the asymmetry in translational drag, 
the corresponding effects cannot be periodically averaged into almost zero, which can provide ventral-dorsal forces and body pitch torques.
\begin{figure}[t]
\centering
\includegraphics[width=2.7in]{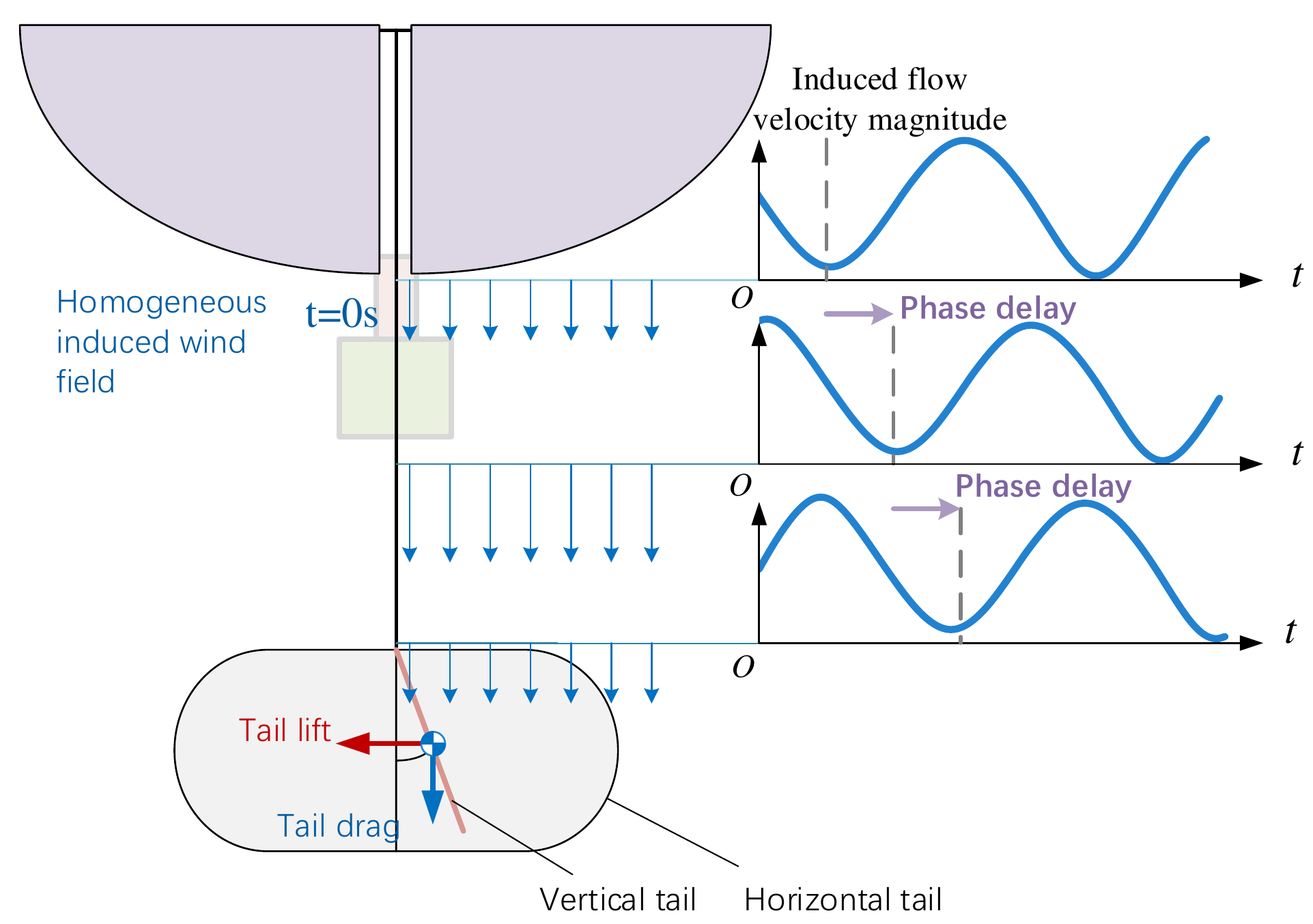}
\caption{Schematic of the homogeneous flapping wing induced wind field and its phase delay. The dissipations of these induced velocities are neglected since the traveling distance from the trailing edge to the tail is relatively small.}
\label{figure:Show1}
\end{figure}
\subsection{Wing-Tail Interaction}
According to our simulation based on the formula \eqref{eq:enduced}, as well as PIV test reported in \cite{Armanini-2019}, the flapping wing induced velocity has a comparable magnitude as the free-stream velocity does. 
With respect to our flapping wing robot, and in usual robotic applications, the maximum magnitude of the induced velocity is $2\mbox{-}4~{\rm m/s}$, and the average distance between the flapping edge and the tail is approximately $15~{\rm cm}$, which indicates that the wake traveling delay
is about $30\mbox{-}80~{\rm ms}$. Compared with the $50\mbox{-}150~{\rm ms}$ flapping wing period length, there is no obvious time scale separation, 
which induces that the phase delay between the flapping wing aerodynamics and the tail aerodynamics is non-negligible. 
And based on experimental results provided in \cite{Armanini-2019}, there is no prominent spanwise phase delay or magnitude difference, if the distance between the wing trailing edge and the tail is not extremely small. 
To this end, we use the integrated resultant lift to compute the induced velocity and suppose the wind field is homogeneous within the range of the actuator disk, instead of computing the induced velocity in each individual blade element and floundering in complex wind fields. 
In this simulation, the induced wind field moves synchronously and caudally within the periodic average speed, where the period is determined by the current flapping wing frequency, which is demonstrated in Fig. \ref{figure:Show1}.
Consequently, the current induced velocity actuated on the tail can be obtained by tracing back previous computed flapping wing induced velocity at a specific time point, in order to simulate the phase delay. 
The reason that we do not set a constant phase delay is the variation of the flapping wing frequency.
However, we cannot record indefinitely, such that we set a record truncation with its capacity of $1000~{\rm ms}$ record length.
If the delay is too large, meaning that the induced velocity is considerably small, then its effect on the rudder can be neglected. 
Following the above settings, we can simulate the frequency-varying wing-tail interactions, even with changing flapping frequencies and in different free flows.

\begin{figure}[t]
\centering
\includegraphics[width=2.7in]{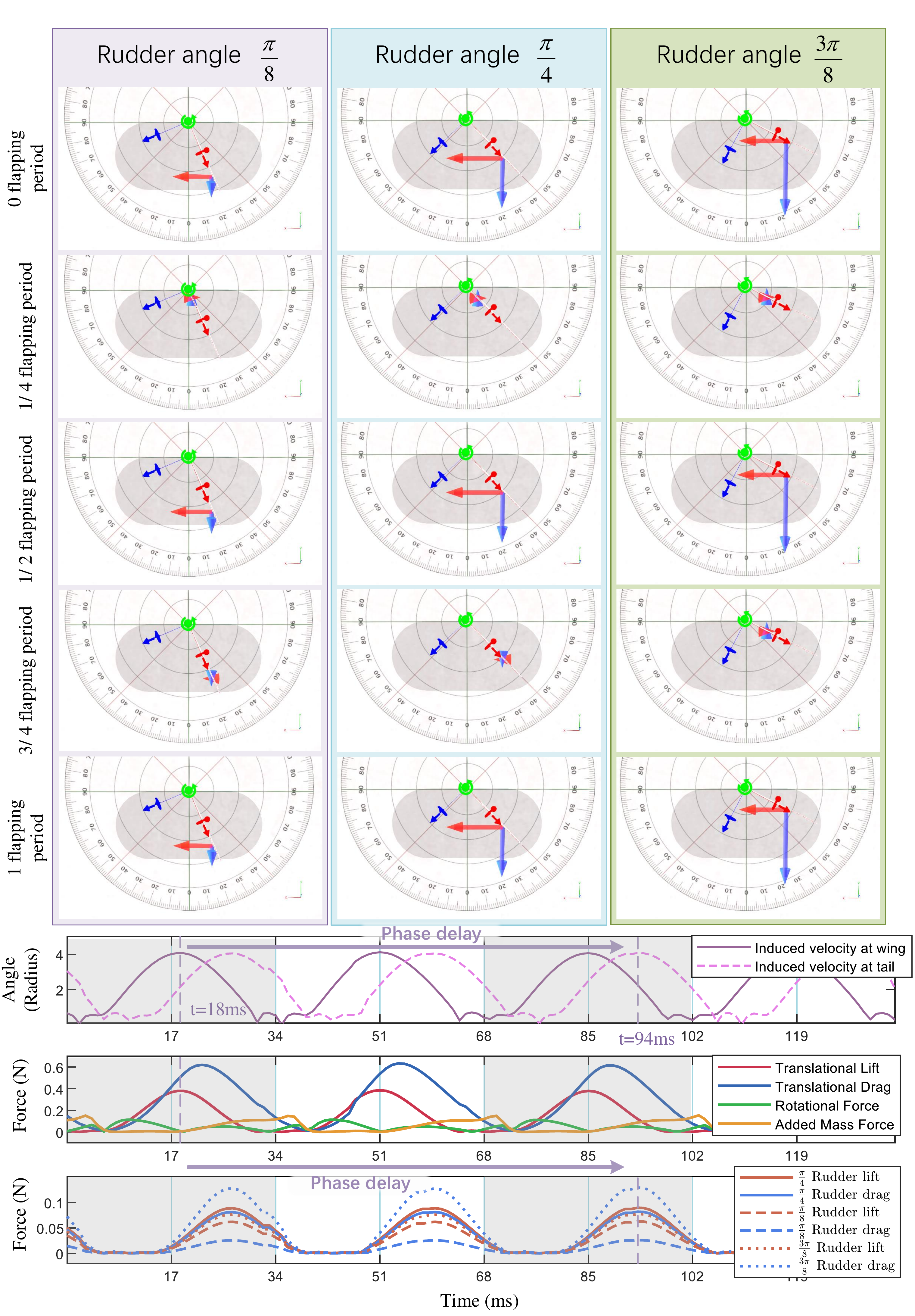}
\caption{Periodic aerodynamic forces actuated on the vertical tail with different vertical tail rudder rotation angles. The frame shown is fixed to the vertical tail, whose green axis coincides with the vertical tail rotation axis. The torsional spring constant is set as $0.025~{\rm N\cdot m/rad}$, stroke frequency at $15{~\rm Hz}$, and the rudder rotation angle is set as $\pi / 8$, $\pi / 4$, and $3 \pi / 8$, respectively.  }
\label{figure:Show2}
\end{figure}

\subsubsection{Periodic tail aerodynamic forces}
The aerodynamic forces of the vertical tail with only the flapping wing induced flow as its incident flow are periodic.
In the following simulation case, we only show the changing forces situations when they are periodically stable, 
which takes several stroke cycles after the wings start to flap.
As shown in the Fig. \ref{figure:Show2}, due to the symmetry of front and back stroke motion, 
the aerodynamic forces actuated on the vertical tail changes twice the frequency of the stroke.
Similar to the wing aerodynamics, the tail lift reaches its peak when its rotation angle is $\pi / {4}$, 
which can also be seen as the aerodynamic AoA in this case. 
The oscillating forces due to the flapping wing induced incident indicate that,
in low speed flight, or hovering flight, the torques generated by the tail are relatively unstable, 
however, can still achieve efficient maneuver. 
When the torsional spring constant is set as $0.025~{\rm N\cdot m/rad}$, stroke frequency at $15{~\rm Hz}$, 
the induced velocity time delay between the wing trailing edge and tail is approximately $76{~\rm ms}$,
or approximately $2.269 \pi$ long phase delay.
The phase delay ${\Phi _{\rm{d}}}$ is computed by 
\begin{align}
\label{eq:dphase}
{\Phi _{\rm{d}}} = 2\pi \frac{{{T_{\rm{d}}}}}{{{T_{{\rm{fp}}}}}}
\end{align}
where $T_{{\rm{d}}}$ is the time delay, and $T_{{\rm{fp}}}$ is the flapping period, equaling to the induced velocity period.

\begin{figure}
\centering
\includegraphics[width=2.9in]{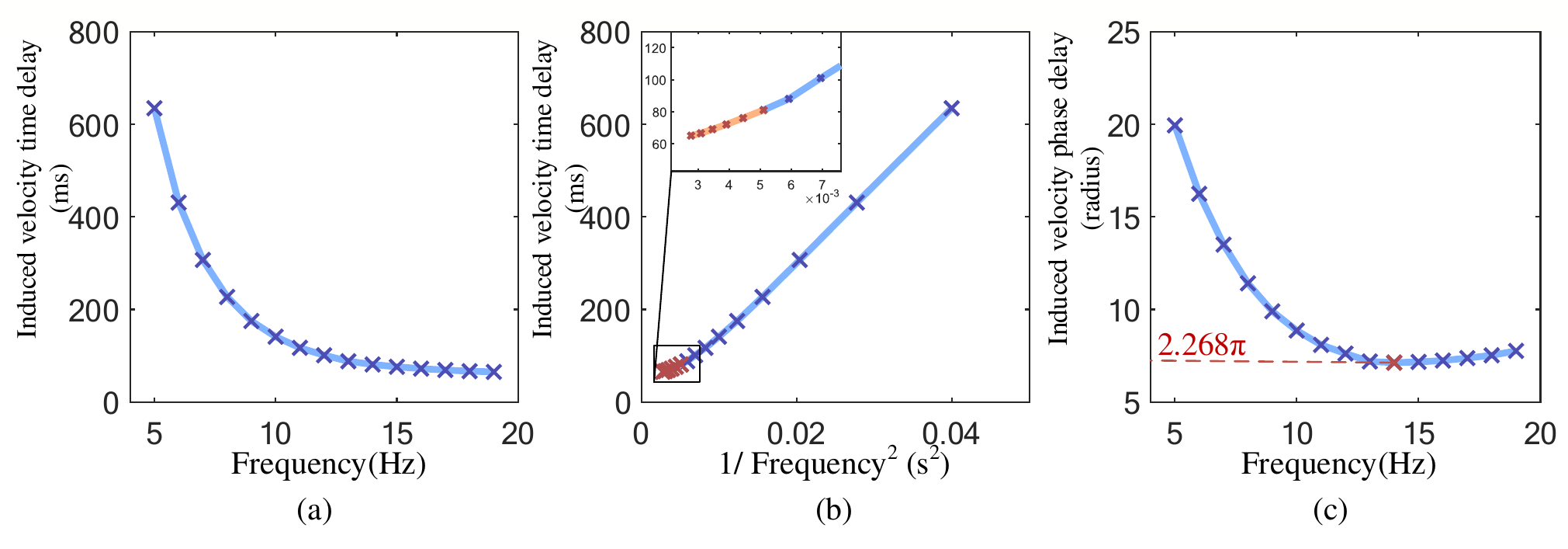}
\caption{Flapping wing induced wake traveling time delay and phase delay from wing trailing edge to tail.  }
\label{figure:DelayAnalysis}
\end{figure}

\subsubsection{Induced velocity delay}
Furthermore, the relationship between flapping frequencies and delays is analyzed.
In this series of simulations, the flapping wing frequency is tested from $5~{\rm Hz}$ to $19~{\rm Hz}$ with $1~{\rm Hz}$ interval, 
which is the frequently used frequency in most applications of the simulated robot.
Based on the simulation result in Fig. \ref{figure:DelayAnalysis}~(a), the induced velocity time delay descends, as the testing flapping frequency increases, while the decreasing rate decreases.
For example, from $5~{\rm Hz}$ to $6~{\rm Hz}$, there is a $204~{\rm ms}$ induced velocity time delay drop, meanwhile, from $18~{\rm Hz}$ to $19~{\rm Hz}$, the drop is only $1~{\rm ms}$. 
This indicates that, with respect to the simulated robot, the induced flow has its upper limit, albeit the flapping wing frequency can still increase.
As shown in Fig. \ref{figure:DelayAnalysis}~(b), there is a strong proportional relationship between induced velocity time delay
and the reciprocal of frequency square, although when the frequency is larger than $14~{\rm Hz}$,
 this proportional relationship gradually attenuates. 
Because the coherence between the flapping wing forces and tail generated torques can provide a relatively periodically stable pattern,
which cannot be completely captured by the averaged dynamics model,
the phase delay is more important in determining the overall robot dynamics and locomotions.
The phase delay versus flapping frequency is given in Fig. \ref{figure:DelayAnalysis}~(c).
Based on our observation, there is no obvious strong relationship between flapping wing frequency or its variants.
The phase delay sharply decreases from $5~{\rm Hz}$ to $10~{\rm Hz}$, then the decrease slows down, and reaches its nadir $2.268 \pi$ at $14~{\rm Hz}$, and then slowly increases.

\subsection{Aerodynamic Forces and Torques Statistics}
Averaging method can capture basic flapping wing robot aerodynamics in relatively stable flight.
In the meantime, from a macro perspective of body dynamics, the fluctuations or oscillations components can be neglected to a considerable extent.  
To facilitate practical robot application, forces and torques macro effects are analyzed within the overall robot, instead of individual wing or deflection surfaces.

\begin{figure*}[t]
      \centering
      \includegraphics[width=6in]{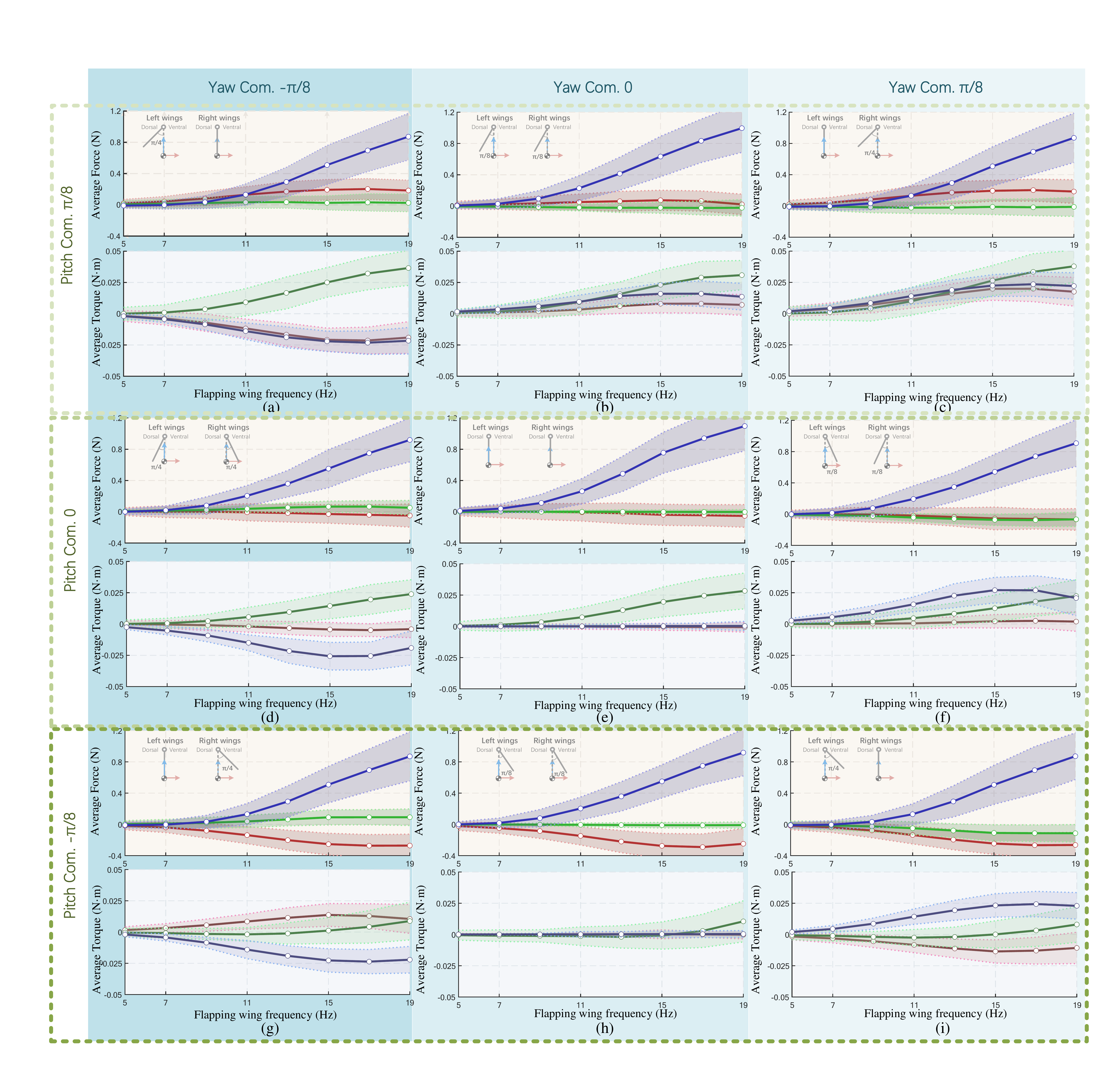}\\
      \caption{Resultant average forces and torques by flapping wings with different actuation settings. 
      The left and right wings equilibrium pitch positions are adjusted to generate different torques.
      Detail wings configurations are sketched at the top left corner in each sub-figure.
      The forces and torques are all resolved in the body-fixed frame, right-handed system.
      Forces and torques along X-axis (ventral) are shown as red curves, Y-axis (left) green, Z-axis (cephalad) blue.
      The shaded area width represents the oscillations that the periodic forces and torques exerted on the robot, $\upsilon$ in \eqref{eq:oscillation}, which is discussed in Appendix A.}
      \label{figure:AveShow}
\end{figure*}

\begin{table}[t]
\caption{Flapping wings actuation allocations}
  \centering
  \label{tab:FWAS}
\begin{tabular}{cccccc}
\toprule
Fig. show &Pitch Com.  &Yaw  Com. &Left Wings  & Right Wings \\
\midrule
{\ref{figure:AveShow}-(a)}&  ${{ \pi } \mathord{\left/{\vphantom {{ \pi } 8}} \right.\kern-\nulldelimiterspace} 8}$& ${{ -\pi } \mathord{\left/{\vphantom {{ -\pi } 8}} \right.\kern-\nulldelimiterspace} 8}$ & Dorsal ${{  \pi } \mathord{\left/{\vphantom {{  \pi } 4}} \right.\kern-\nulldelimiterspace} 4}$ & Neutral\\
{\ref{figure:AveShow}-(b)}&  ${{ \pi } \mathord{\left/{\vphantom {{ \pi } 8}} \right.\kern-\nulldelimiterspace} 8}$& 0 & Dorsal ${{  \pi } \mathord{\left/{\vphantom {{  \pi } 8b}} \right.\kern-\nulldelimiterspace} 8}$ & Dorsal ${{  \pi } \mathord{\left/{\vphantom {{  \pi } 8}} \right.\kern-\nulldelimiterspace} 8}$\\
{\ref{figure:AveShow}-(c)}&  ${{ \pi } \mathord{\left/{\vphantom {{ \pi } 8}} \right.\kern-\nulldelimiterspace} 8}$& ${{ \pi } \mathord{\left/{\vphantom {{ \pi } 8}} \right.\kern-\nulldelimiterspace} 8}$ & Neutral & Dorsal ${{  \pi } \mathord{\left/{\vphantom {{  \pi } 4}} \right.\kern-\nulldelimiterspace} 4}$\\
{\ref{figure:AveShow}-(d)}&  0& ${{ -\pi } \mathord{\left/{\vphantom {{ -\pi } 8}} \right.\kern-\nulldelimiterspace} 8}$ & Dorsal ${{  \pi } \mathord{\left/{\vphantom {{  \pi } 8}} \right.\kern-\nulldelimiterspace} 8}$ & Ventral ${{  \pi } \mathord{\left/{\vphantom {{  \pi } 8}} \right.\kern-\nulldelimiterspace} 8}$\\
{\ref{figure:AveShow}-(e)}&  0& 0 & Neutral & Neutral\\
{\ref{figure:AveShow}-(f)}&  0& ${{ \pi } \mathord{\left/{\vphantom {{ \pi } 8}} \right.\kern-\nulldelimiterspace} 8}$ &  Ventral ${{  \pi } \mathord{\left/{\vphantom {{  \pi } 8}} \right.\kern-\nulldelimiterspace} 8}$ &  Dorsal ${{  \pi } \mathord{\left/{\vphantom {{  \pi } 8}} \right.\kern-\nulldelimiterspace} 8}$\\
{\ref{figure:AveShow}-(g)}&  ${{ -\pi } \mathord{\left/{\vphantom {{ -\pi } 8}} \right.\kern-\nulldelimiterspace} 8}$& ${{ -\pi } \mathord{\left/{\vphantom {{ -\pi } 8}} \right.\kern-\nulldelimiterspace} 8}$& Neutral &  Ventral ${{  \pi } \mathord{\left/{\vphantom {{  \pi } 4}} \right.\kern-\nulldelimiterspace} 4}$\\
{\ref{figure:AveShow}-(h)}&  ${{ -\pi } \mathord{\left/{\vphantom {{ -\pi } 8}} \right.\kern-\nulldelimiterspace} 8}$& 0 & Ventral ${{  \pi } \mathord{\left/{\vphantom {{  \pi } 8}} \right.\kern-\nulldelimiterspace} 8}$ & Ventral ${{  \pi } \mathord{\left/{\vphantom {{  \pi } 8}} \right.\kern-\nulldelimiterspace} 8}$\\
{\ref{figure:AveShow}-(i)}&  ${{ -\pi } \mathord{\left/{\vphantom {{ -\pi } 8}} \right.\kern-\nulldelimiterspace} 8}$& ${{ \pi } \mathord{\left/{\vphantom {{ \pi } 8}} \right.\kern-\nulldelimiterspace} 8}$ & Ventral ${{  \pi } \mathord{\left/{\vphantom {{  \pi } 4}} \right.\kern-\nulldelimiterspace} 4}$ & Neutral\\
\bottomrule
\end{tabular}
\end{table}

\subsubsection{Flapping wing dynamics effects}
Averaging method can be applied to flapping wing robot relatively stable flight. 
Investigating average forces and torques induced by different flapping wings actuation settings 
can provide fundamental knowledge for robot actuation manner.   
Furthermore, the oscillation inputs can significantly change the robot overall behaviors,
such that we have to investigate and quantify them.
To this end, we propose a statistic based on dynamic differences from the nominal averaged system:
\begin{equation}
\upsilon  = {\rm sqrt}(\mathop {\sup }\limits_{\scriptstyle{\tau _1},{\tau _2} \in \left[ {{t_0},{t_1}} \right]\hfill\atop
\scriptstyle{\tau _1} < {\tau _2}\hfill} \left\| {\sum\limits_{s = {\tau _1}}^{{\tau _2}} {\left( { f\left( s \right) - \bar { f}} \right)} }\cdot l \right\|)
\label{eq:oscillation}
\end{equation}
for a continuous sampling interval from $t_0$ to $t_1$, 
where ${\rm sqrt}( \star ): \mathbb R \to \mathbb R$ is the square root function, $\bar {f} = {1 \mathord{\left/
 {\vphantom {1 {\left( {{t_1} - {t_0}} \right)}}} \right.
 \kern-\nulldelimiterspace} {\left( {{t_1} - {t_0}} \right)}} \cdot \sum\nolimits_{s = {t_0}}^{{t_1}} {f\left( s \right)} $, $l$ is the simulation step, and $f$ can be selected from the flapping wing forces or torques, which provides a unified oscillation description for flapping wing dynamics under different flapping frequency based on the effects of momentum inputs. 
The development of this statistic is elaborated in Appendix A.
It is noteworthy that only the relative magnitude of $\upsilon$ shows practical significance. And since the moment of inertia is not considered in $\upsilon$, 
this statistic describes the effects of the time-varying force or torque, rather than its consequent kinematic behavior.

When we set the wing pitch torsional spring constant as $0.025~{\rm N\cdot m/rad}$, and stroke frequency as $5$-$19{~\rm Hz}$ with $2{~\rm Hz}$ interval, and also fix the robot with no translational or angular velocity,
we obtain the simulation result shown in Fig.~\ref{figure:AveShow}. The detailed map between torque command and wings pitch balance position settings, is given in TABLE \ref{tab:FWAS}.
Due to the intrinsic unsteadiness and oscillation in flapping wing motions, the force and torques corresponding to the commands, such as pitch command or yaw command, are highly nonlinear and coupled.
We list the following four notable phenomena of wing deviated from expectation: 
\begin{enumerate}
  \item[ $\bullet$ ] PW-1: (\emph{Torque-generating inducing thrust drop}) Comparing obtained results in Fig.~\ref{figure:AveShow}-b,d,f,h with that in Fig.~\ref{figure:AveShow}-e, respectively, we can find that, when the robot tries to generate pitch or yaw torque with ${\pi  \mathord{\left/{\vphantom {\pi  8}} \right.\kern-\nulldelimiterspace} 8}$ wings balance positions changing, there emerges an approximate $23\%$ drop of the thrust.
  \item[ $\bullet$ ] PW-2: (\emph{Intrinsic pitch forward}) Simulation tests corresponding to Fig.~\ref{figure:AveShow}-a,b,c,d,f,g,h,i, when the flapping wing frequency increases, the generated pitch and yaw torques first increase simultaneously, however, start to decrease after 17 Hz, which is highly nonlinear.
  Meanwhile, the pitch torque increases approximately proportionally to flapping wing frequency, even without a pitch command, because the center of the stroke plane is dorsally displaced from the mass center.
  \item[ $\bullet$ ] PW-3: (\emph{Wings asymmetry generating roll torque}) Investigating the test group Fig.~\ref{figure:AveShow}-a,d,g, and the group Fig.~\ref{figure:AveShow}-c,f,i, we can find that, when the rotation magnitude of left and right wings pitch balance position is different, the unexpected roll torques arise, which manifests the high coupling characteristic.
  \item[ $\bullet$ ] PW-4: (\emph{Saturated oscillation}) The oscillation input slowly increases as the flapping wing frequency increases. However, when the frequency passes 15~Hz, the oscillation input magnitude almost remains stagnant thereafter. 
\end{enumerate}

\begin{figure}[t]
\centering
\includegraphics[width=2.7in]{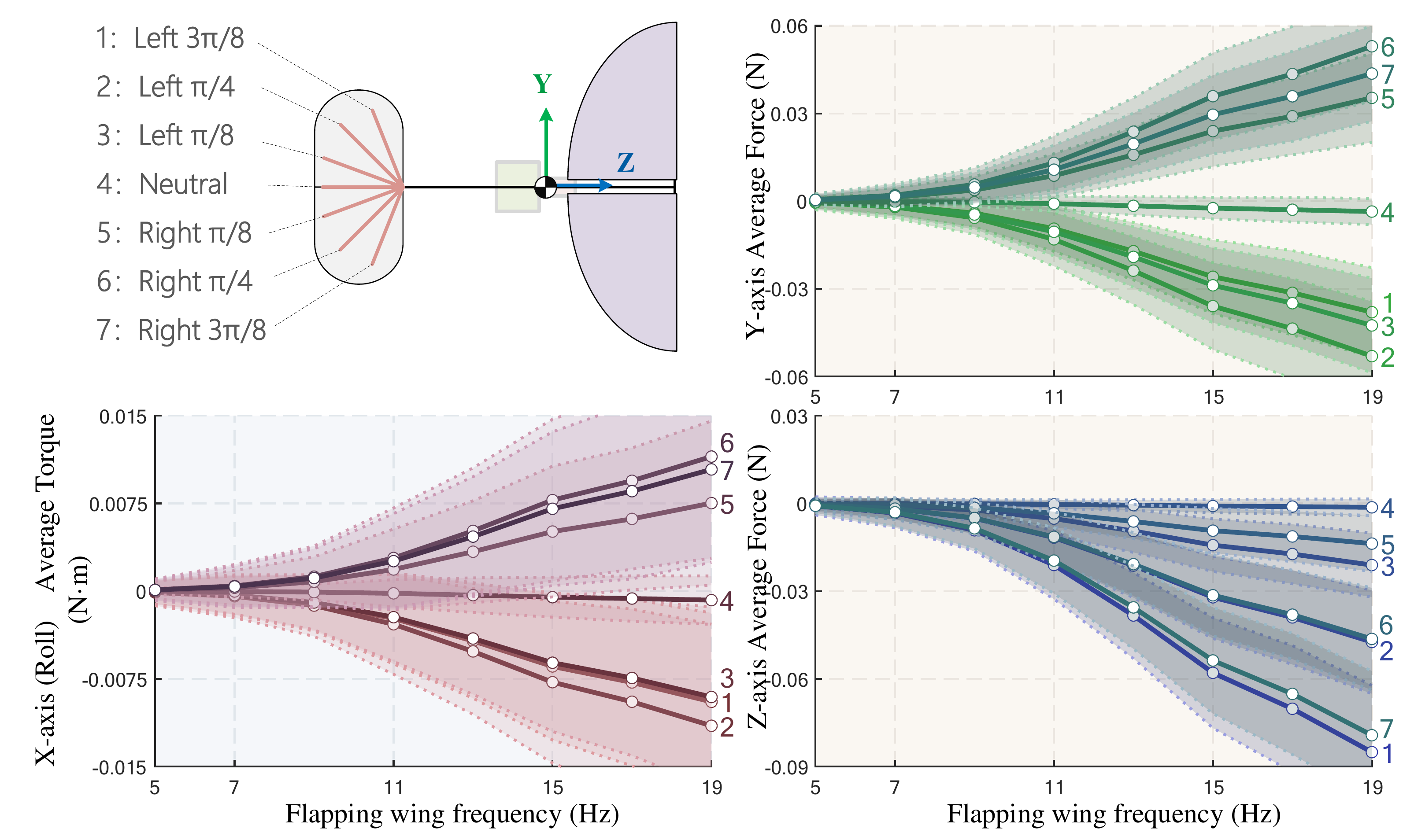}
\caption{Forces and torques induced from different vertical tail rudder rotation angles when the flapping wing torsional spring constant is set as $0.025~{\rm      N\cdot m/rad}$, wing torque command as 0 pitch and 0 yaw. 
The shaded area width represents the oscillations that the periodic forces and torques exerted on the robot.
      The vertical tail rudder rotation angle is adjusted to generate different roll torques.
      The specific rotations are shown in the left upper corner.
      Forces and torques along X-axis (ventral) are shown as red curves, Y-axis (left) green, Z-axis (cephalad) blue. }
\label{figure:RudderFT}
\end{figure}

\subsubsection{Tail deflection effects} The aforementioned analyses can also be applied to the tail. The vertical tail rotation method is given in TABLE \ref{tab:VTAS}.
In this simulation, we focus on the flapping wings induced incident velocity actuating on the vertical tail, such that
the generated forces and torque are also periodic. 
Based on our observation of the obtained result shown in Fig.~\ref{figure:RudderFT},
the following two phenomena are noteworthy: 

\begin{table}[t]
\caption{Vertical tail rudder actuation allocations}
  \centering
  \label{tab:VTAS}
\begin{tabular}{cccccc}
\toprule
Fig. show &Roll Com.  &Rudder  \\
\midrule
{\ref{figure:RudderFT}-1}&  -${{3 \pi } \mathord{\left/{\vphantom {{3 \pi } 8}} \right.\kern-\nulldelimiterspace} 8}$& Left ${{3 \pi } \mathord{\left/{\vphantom {{ -3 \pi } 8}} \right.\kern-\nulldelimiterspace} 8}$\\
{\ref{figure:RudderFT}-2}&  -${{ \pi } \mathord{\left/{\vphantom {{ \pi } 4}} \right.\kern-\nulldelimiterspace} 4}$& Left ${{ \pi } \mathord{\left/{\vphantom {{ -\pi } 4}} \right.\kern-\nulldelimiterspace} 4}$\\
{\ref{figure:RudderFT}-3}&  -${{ \pi } \mathord{\left/{\vphantom {{ \pi } 8}} \right.\kern-\nulldelimiterspace} 8}$& Left ${{ \pi } \mathord{\left/{\vphantom {{ -\pi } 8}} \right.\kern-\nulldelimiterspace} 8}$  \\
{\ref{figure:RudderFT}-4}&  0& Neutral \\
{\ref{figure:RudderFT}-5}&  ${{ \pi } \mathord{\left/{\vphantom {{ \pi } 8}} \right.\kern-\nulldelimiterspace} 8}$& Right ${{ \pi } \mathord{\left/{\vphantom {{ -\pi } 8}} \right.\kern-\nulldelimiterspace} 8}$  \\
{\ref{figure:RudderFT}-6}&  ${{ \pi } \mathord{\left/{\vphantom {{ \pi } 4}} \right.\kern-\nulldelimiterspace} 4}$& Right ${{ \pi } \mathord{\left/{\vphantom {{ -3\pi } 8}} \right.\kern-\nulldelimiterspace} 4}$ \\
{\ref{figure:RudderFT}-7}&  ${{3 \pi } \mathord{\left/{\vphantom {{3 \pi } 8}} \right.\kern-\nulldelimiterspace} 8}$& Right ${{ 3\pi } \mathord{\left/{\vphantom {{ 3\pi } 8}} \right.\kern-\nulldelimiterspace} 8}$\\
\bottomrule
\end{tabular}
\end{table}

\begin{enumerate}
  \item[ $\bullet$ ] PR-1: (\emph{Roll peaking at $\pi / 4$}) The maximum roll torque appears at the $\pi / 4$ rudder rotation.  Compared to the torque generated by the wings,
  when the rudder completely depends on the flapping wings induced incident flow,
  the roll torque generated by the rudder introduces more oscillations into the system.
  Further, we can conclude that there is a strong proportional relationship between flapping wing frequency between 9-17 Hz.   
  \item[ $\bullet$ ] PR-2: (\emph{Rudder rotation generating Y-axis and Z-axis force}) The coupling between forces and torques is prominent. The rudder induced roll torque substantially depends on the force along Y-axis of the body fixed frame, reflected in the similar patterns shown in Fig.~\ref{figure:RudderFT}, 
  which is due to the salient discrepancy between Y-axis and Z-axis force arms lengths. Furthermore, the deflection of the rudder surface 
  can lead to a maximally $7$-$10\%$ drop in the robot thrust force.
\end{enumerate}

\begin{figure*}[t]
      \centering
      \includegraphics[width=6in]{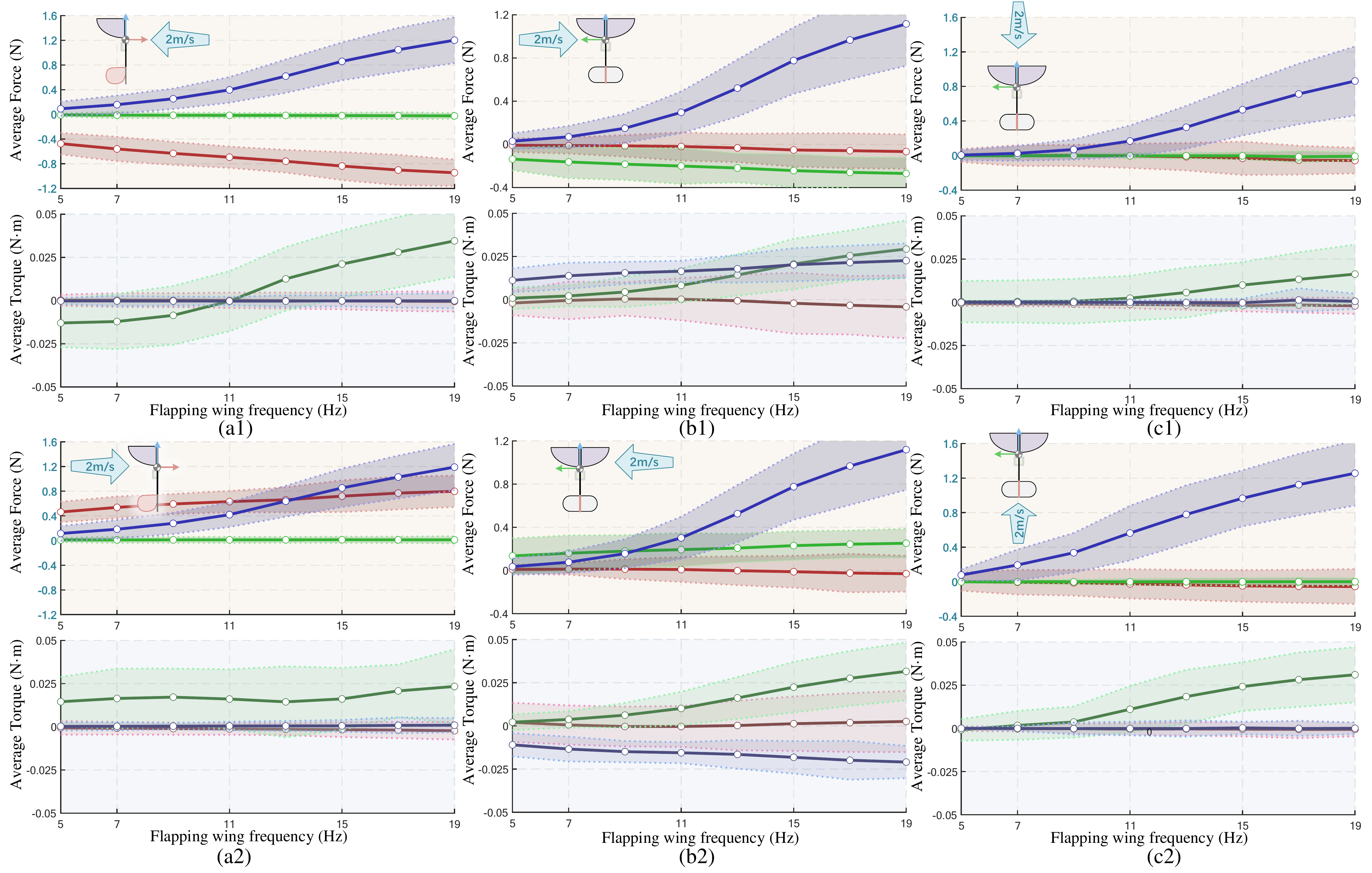}\\
      \caption{Resultant average forces and torques generated by both wings and tails faced with different free flows.
      The shaded area width represents the oscillations that the periodic forces and torques exerted on the robot.
      The wings and tails are all set as neutral positions, corresponding to the situation of Fig. \ref{figure:AveShow}-e and 4 in Fig. \ref{figure:RudderFT}. 
      Detail free flows are sketched at the top left upper corner of each sub-figure.
      The forces and torques are all resolved in the body-fixed frame, right-handed system.
      Forces and torques along X-axis (ventral) are shown as red curves, Y-axis (left) green, Z-axis (cephalad) blue.}
      \label{figure:AveShowVel}
\end{figure*}

\subsubsection{Velocity induced effects}
From the obtained results shown in Fig.~\ref{figure:AveShowVel}, we can investigate the free flow induced dynamics of the robot.
In this series of simulations, we set the robot with 0 pitch and 0 yaw commands, and rudder in neutral position.
First, the windless condition shown in  Fig.~\ref{figure:AveShow}-e can be seen as a static reference, where the tail generates no forces or torques 
such that the robot dynamics completely depends on the wings. Based on our observation of the obtained simulation result shown in Fig.~\ref{figure:AveShowVel},
the following three phenomena of free flow induced aerodynamics can be concluded:

\begin{enumerate}
  \item[ $\bullet$ ] PF-1: (\emph{Resistance force proportional to windward area}) Comparing the pairs (a), (b), and (c) in Fig.~\ref{figure:AveShowVel}, 
  we find that there is obvious resistance force in the opposite direction of free flow velocity,
  whose magnitude is approximately proportional to windward area.
  \item[ $\bullet$ ] PF-2: (\emph{Velocity induced pitch}) In the pairs of simulation (a1) and (a2) in Fig.~\ref{figure:AveShowVel},
   there exists an obvious correlation between  Y-axis (pitch) torque and the free flow velocity along the X-axis.
   The free flow velocity in the negative X-axis direction can induce a negative pitch torque, and vice versa.
   This phenomenon is similar to the \emph{Pendulum-like dynamics} reported in \cite{Pre1}.
  \item[ $\bullet$ ] PF-3: (\emph{Velocity induced roll}) In the pairs of simulation (b1) and (b2) in Fig.~\ref{figure:AveShowVel},
   there also exists an obvious correlation between  Z-axis (yaw) torque and the free flow velocity along the Y-axis.
   The free flow velocity in the negative Y-axis direction can induce a positive pitch torque, and vice versa.
   This phenomenon is similar to the \emph{Wind-vane-like dynamics} reported in \cite{Pre1}.
\end{enumerate}

\section{Simulating Practical Robot Applications}
In the purpose of elaborating that the proposed simulation platform is indeed a flexible, easy-to-use, and application-oriented simulation platform, 
we demonstrate several robotic applications in this section.
In the following simulations, 
all the above aerodynamic force computations are losslessly implemented,
such that we can drill into each subtle behavior in corresponding applications.
Simulations are realized on the Webots \cite{Webots} platform, and programmed mainly in Python.

\subsection{Filtering Oscillations from Flapping Wings}
Before entering into the controllers development section, we first need an attitude filter to reduce the oscillations induced by flapping wings.
Previous methods focus on providing smooth feedback signals and sensor fusion\cite{Wang-2022,Tu-2018}. 
Because the attitude itself is integrated from angular velocity and not excessively oscillating, such that we can straightforwardly use raw attitude data. However, the angular velocity needs filtering to reduce the large-scale oscillation, where a low-pass second order filter is implemented in each channel. 
The corresponding transfer function is given by
\begin{equation}
H\left( s \right) = \frac{{\omega _n^2}}{{{s^2} + 2\zeta {\omega _n}s + \omega _n^2}}
\end{equation}
where $\omega_n \in \mathbb R $ is the natural frequency, satisfying $\omega_n = 2 \pi f_n$, and $\zeta\in \mathbb R $ is the damping ratio.
Since the flapping frequency in flight is approximately from 9Hz to 15Hz, we set  $\omega_n = 2\pi \cdot 8 $, and chose $\zeta = 0.8$ according to 
the simulation performance. 

\begin{figure}[t]
      \centering
      \includegraphics[width=3.2in]{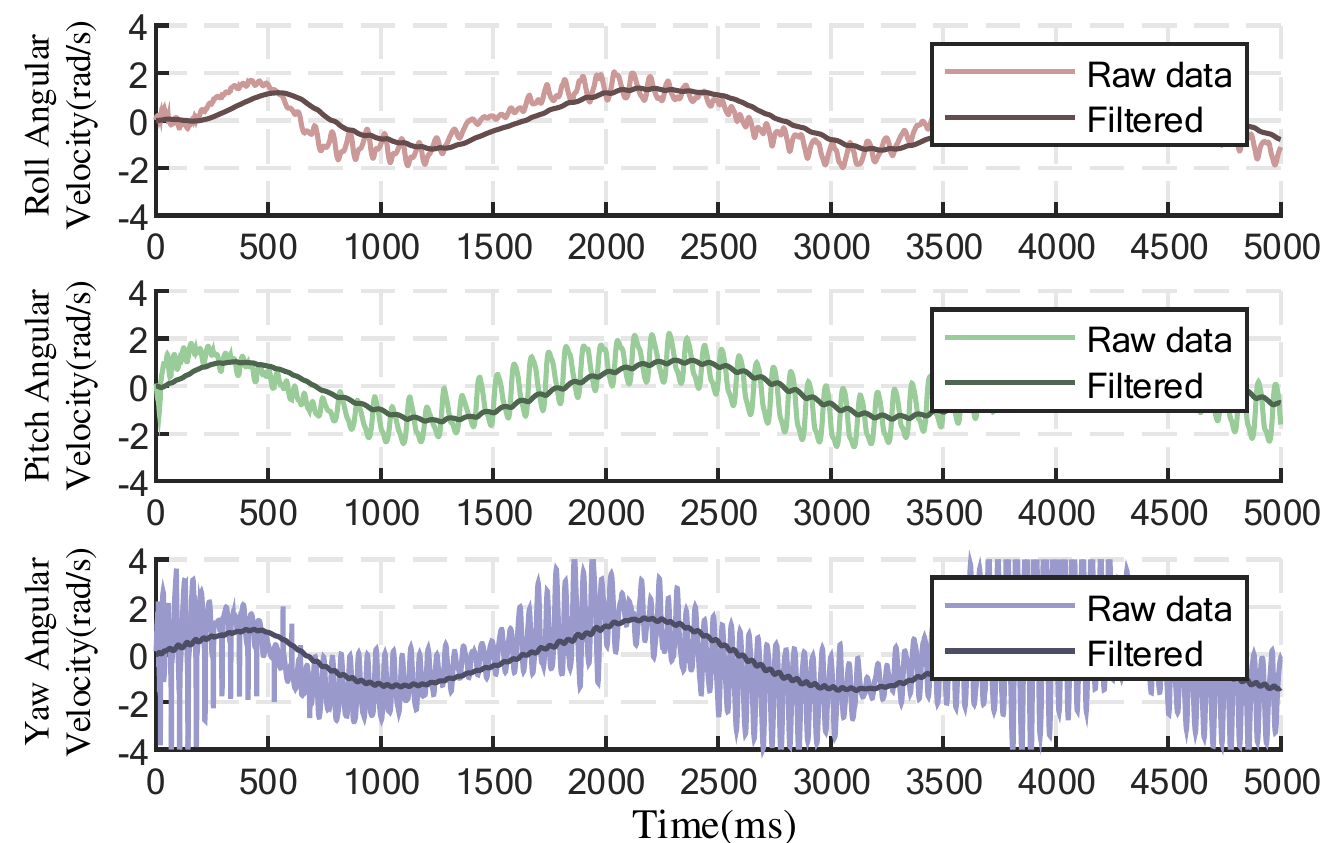}\\
      \caption{Low-pass second order filter processing oscillating angular velocity signal in simulation.}
      \label{figure:FilterShow}
\end{figure}

As shown in Fig. \ref{figure:FilterShow}, the oscillation induced by the flapping wings can be filtered by the designed filter.
According to our observation in both simulations and experiments,
implementing the corresponding filtered signal as the controller feedback can modify the robot flight performance.

\subsection{Attitude Tracking}
Stable attitude tracking is one of the prerequisites for deploying other flapping wing robot tasks.
The basic idea of our developed attitude tracking controller originates in \cite{Lee-2013}. 
Implementing averaging method, 
the flapping wing attitude controller is to solve the rigid body attitude tracking problem on ${\rm SO(3)}$, where the  
attitude representation complexities and ambiguities can be avoided. 
The ${\rm SO(3)}$ group is 
\begin{equation}
\label{eq:SO3group}
{\rm{SO}}\left( 3 \right) = \left\{ {R \in \mathbb R^{3 \times 3}}\left| {R{R^{\top}} = {\bm I_3},\det \left[ R \right] = 1} \right. \right\},
\end{equation}
where $\det[\star]$ is the determinant of $\star$. The Lie algebra associated with ${\rm{SO}}\left( 3 \right)$ is ${\mathfrak{so}}\left( 3 \right) = \{ {S \in {\mathbb R^{3 \times 3}}\left| {S =  - {S^\top}} \right.} \}$.

The control objective is to make the robot orientation following the desired attitude trajectory, which is shown as
\begin{equation}
\label{eq:Desire}
{\dot R_d} = {R_d}{\hat {\bm \Omega} _d}
\end{equation}
where ${\bm \Omega}_d \in \mathbb R^3$ is the desired angular rotation velocity, 
$R_d \in \mathbb {\rm SO(3)} $ is the desired rotation matrix. 

First of all, the attitude dynamics is 
\begin{align}
J\dot {\bm \Omega}    &= {\bm \tau}  + \delta {\bm \tau}  - {\bm \Omega}   \times J{\bm \Omega}    \\
\dot R &= R\hat {\bm \Omega}   
\end{align}
where $J \in \mathbb R^{3 \times 3} $ is the inertia matrix in the body fixed frame, the hat map $\hat{\bm \star} : \mathbb R^3 \to {\mathfrak{so}}\left( 3 \right)$ maps an angular velocity vector to a skew symmetric matrix, such that $\hat{\bm \star}_1\bm \star_2=\bm \star_1 \times \bm \star_2$,  ${\bm \Omega} \in \mathbb R^3$ is the angular rotation velocity, $R \in \mathbb {\rm SO(3)} $ is the rotation matrix representing the robot attitude in the inertia frame,
$\bm \tau \in \mathbb R^3$ is the torque input, 
and $\delta \bm \tau \in \mathbb R^3$ is the composition of the torque disturbances and actuator faults from model uncertainties and actuator misalignment.

The attitude error function is 
\begin{equation}
\label{eq:ErrorFunc}
\Psi \left( {R,{R_d}} \right) = \frac{1}{2}{\rm{tr}}\left[ {G\left( {I - R_d ^\top R} \right)} \right]
\end{equation}
where $R_d \in \mathbb {\rm SO(3)}$ is the desired robot orientation in the inertia frame, ${\rm tr}(\star)$ indicates the trace of $\star$, and 
$G \in \mathbb R^{3 \times 3}$ is a positive definite, diagonal matrix. 
The attitude error $\Psi \left( {R,{R_d}} \right)$ dynamics is shown as 
\begin{align}
\frac{{\rm{d}}}{{{\rm{d}}t}}\left( {\Psi \left( {R,{R_d}} \right)} \right) &= \bm e_R^\top{\bm e_\Omega },\\
{\bm e_R} &= \frac{1}{2}{\left( {GR_d^\top R - {R^\top}{R_d}G} \right)^ \vee },\\
{\bm e_\Omega } &=\bm \Omega  - {R^\top}{R_d}{\bm \Omega _d},\label{eq:ErrorOmega}\\
{{\dot {\bm e}}_R} &= \frac{1}{2}{\left( {GR_d^\top R{{\hat {\bm e}}_\Omega } + {{\hat {\bm e}}_\Omega }{R^\top}{R_d}G} \right)^ \vee },\\
{{\dot {\bm e}}_\Omega } &= {J^{ - 1}}\left( { - \bm \Omega  \times J\bm \Omega  + \bm  u + \delta \bm \tau } \right) - {\bm \alpha _d}
\end{align}
where $(\star)^\vee : {\mathfrak{so}}\left( 3 \right) \to \mathbb R^3 $ is the inverse map of $\hat{\star}$, ${\bm \alpha _d} \in \mathbb R^3$ is the desired angular acceleration terms composition in the current body-fixed frame, which is given by
\begin{equation}
{\bm \alpha _d} =  - \hat {\bm \Omega} {R^\top}{R_d}{{\bm \Omega} _d} + {R^\top}{R_d}{\dot {\bm \Omega} _d}
\end{equation}

\begin{Remark}
\label{Re:not_hybrid}
The attitude error we use in this simulation has three unstable non-trivial critical points, and begin to decays after traveling $\pi / 2$ along each curve corresponding to the geodesic spray at the initial point, which makes it less efficient in practical application. Hybrid attitude errors\cite{Pre1,Lee-2015} although with stronger system stability convergences in theory, however, owing to their partial discrete nature, will inevitably infuse detrimental non-smooth dynamic factors, which prevents these methods from being invoked in flapping wing robots usually requiring a relatively steady airflow condition.    
\end{Remark}

\begin{Remark}
\label{Re:flapp_att_pro}
Besides issues in conventional 6-DoF attitude control problems, the attitude tracking control problem of the flapping wing robot has the following two challenging aspects: 
\begin{enumerate}
\item  How to generate accurate $\bm \tau$ which is strongly influenced by the flying state.
\item  How to reject the undesired $\delta \bm \tau$ disturbances as much as possible, by practical inputs and actuators.
\end{enumerate}
\end{Remark}

There exist non-negligible differences between different robots and flight situations, such that exhaustive investigation of those maps is nearly impossible. Therefore we use the model reference adaptive method and robust controller design to attack this problem, which makes the controller more robust.

By synthesizing the attitude error and the angular velocity error, and combining them with adaptive and robust terms,
the control torque $\bm \tau$ is given by
\begin{equation}
\label{eq:controller}
\bm \tau  = \underbrace { - {k_R}{\bm e_R} - {k_\Omega }{\bm e_\Omega } + \bm v}_{{\rm{Feedback~terms}}} + \underbrace {\bm \Omega  \times \bar J \bm \Omega  + \bar J{\bm \alpha _d}}_{{\rm{Feedfoward~terms}}}
\end{equation}
where $k_R, k_\Omega \in \mathbb R$ are positive constants, $\bm e_R, \bm e_\Omega \in \mathbb R^3$ are attitude error and angular velocity error, 
respectively, $\bm v \in \mathbb R^3$ is the robust term, $\bar J \in \mathbb R^{3 \times 3} $ is the estimated inertia matrix.
Since the inertia matrix usually can be approximately estimated, the update of the $\bar J$ can be given by
\begin{align}
\label{eq:Jestimation}
\dot {\bar J} = &\frac{{{k_J}}}{2}\left( { - {\bm \alpha _d}\bm e_A^{\top} - {\bm e_A}\bm \alpha _d^{\top} + \bm \Omega {\bm\Omega ^{\top}}{{\hat {\bm e}}_A} - {{\hat {\bm e}}_A}\bm \Omega {\bm \Omega ^{\top}}} - 2 \sigma \bar J \right) \nonumber \\
\end{align}
where $\bm e _A$ is the composition of attitude error and angular velocity error ${\bm e_A} = {\bm e_\Omega } + c{\bm e_R}$, and $\sigma \in \mathbb R$ is the update damping rate.
Inspired by the attitude control scheme reported in \cite{Lee-2013, Lu-2013}, 
the robust term $\bm v$ can be designed in the following two ways, which are high frequency control, and synthesization of high frequency control and sliding mode control, respectively:
\begin{align}
\bm v_1 &=  - \frac{{\delta { _b}^2{\bm  e_A}}}{{\delta { _b}\left\| {{\bm e_A}} \right\| + \varepsilon }}\\
\bm v_2 &=  - {k_v}{\textbf{{sgn}}} \left( {{\bm  e_A}} \right){\left\| {{\bm  e_A}} \right\|^{\rho_v }} - \frac{{\delta { _b}^2{\bm  e_A}}}{{\delta { _b}\left\| {{\bm e_A}} \right\| + \varepsilon }}
\end{align}
where $k_v, 0<\rho_v<1 \in \mathbb R$ are positive constants, $\textbf { sgn}(\star) : \mathbb R ^3 \to \mathbb R ^3$ is the signal function which returns the signal of each element in the vector, which returns $-1$ for negative input, and $1$ for positive, $\delta { _b} \in \mathbb R  $ is the module bound of $\delta \bm \tau$, and $\varepsilon \in \mathbb R$ is a relatively small constant to avoid singularity.


From phenomena PW-1, PR-2, and PF-1,
we conclude that robot resultant force and torque are not only influenced by the flapping frequency 
but many other control commands and flying states, such that, even only along body-fixed frame Z-axis, 
 the controller performance is not consistent
at least under different flapping wing frequencies and desired attitudes. 
In the following attitude tracking control simulation, we use the actuation allocations provided in TABLE~\ref{tab:FWAS} and TABLE~\ref{tab:VTAS}, where the roll command is constrained within $\left[ { - {\pi  \mathord{\left/
 {\vphantom {\pi  4}} \right.
 \kern-\nulldelimiterspace} 4},{\pi  \mathord{\left/
 {\vphantom {\pi  4}} \right.
 \kern-\nulldelimiterspace} 4}} \right]$ to maintain the monotonicity.
 Similarly, equilibrium pitch positions are also restricted.    
 Applying the aforementioned attitude tracking controller, the challenges highlighted in \emph{Remark \ref{Re:flapp_att_pro}} can be basically overcome\footnote{See source codes at https://github.com/Chainplain/Flapping$\_$wing$\_$Simu.}. 
 However, in order to search a better attitude tracking controller, derived from \eqref{eq:controller}, the following three controllers are implemented and further compared:
\begin{align}
{\bm \tau _1} &=  - {k_R}{\bm e_R} - {k_\Omega }{\bm e_\Omega }  \nonumber\\
{\bm \tau _2} &=  - {k_R}{\bm e_R} - {k_\Omega }{\bm e_\Omega } + {\bm v_1} + \bm \Omega  \times \bar J{\bm \Omega}  + \bar J{\bm \alpha _d} \nonumber\\
{\bm \tau _3} &=  - {k_R}{\bm e_R} - {k_\Omega }{\bm e_\Omega } + {\bm v_2} \label{eq:pro_at_con}
\end{align}
Basically, the controller ${\bm \tau _1}$ is similar to a proportional differential (PD) controller, the controller ${\bm \tau _2}$
is a robust adaptive controller, and controller ${\bm \tau _3}$
is a robust sliding mode controller.
They all use the same attitude error function.
\begin{figure*}[t]
      \centering
      \includegraphics[width=6in]{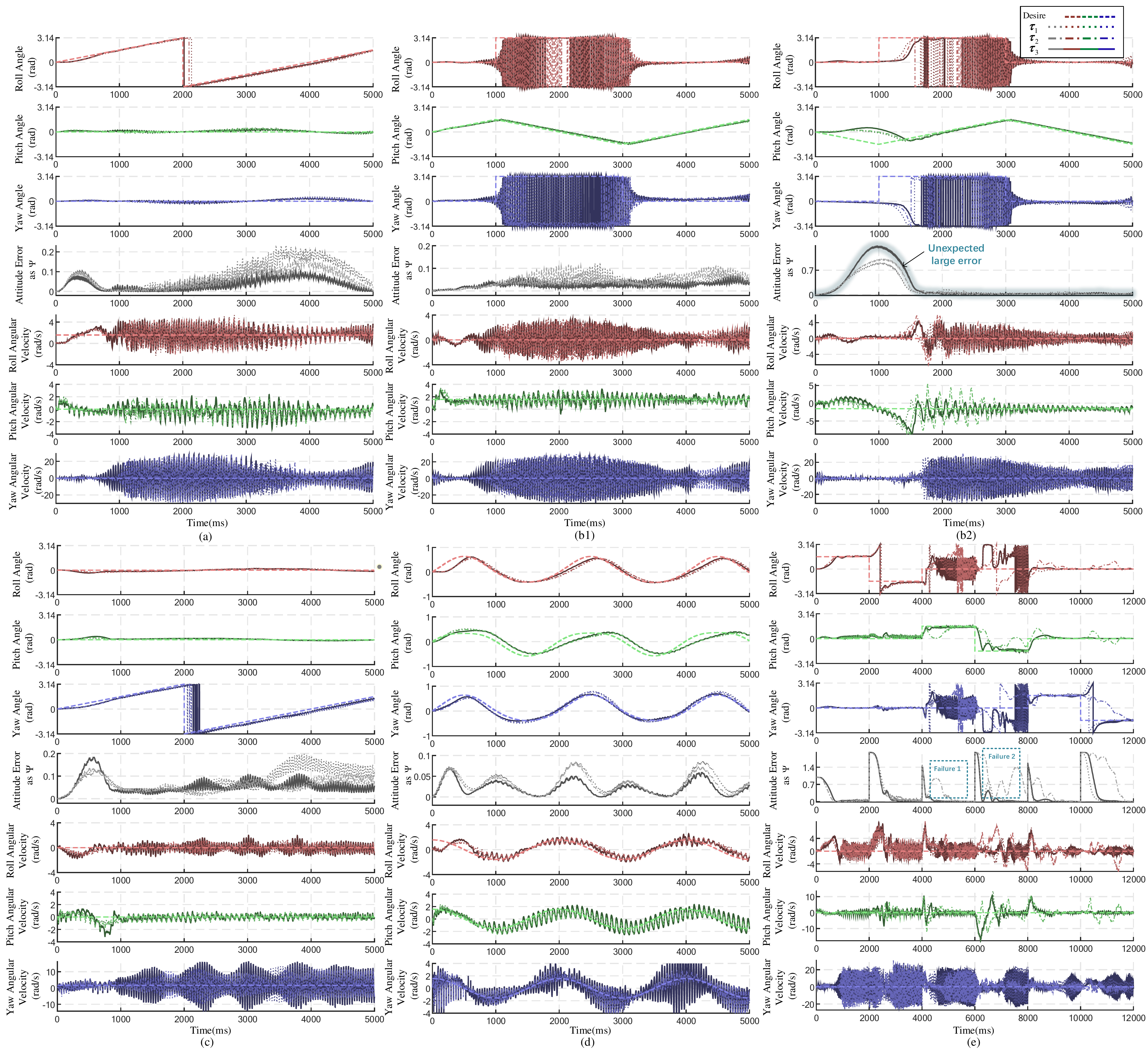}\\
      \caption{Attitude tracking control simulations for three different control strategies, $\bm \tau_1$, $\bm \tau_2$, and $\bm \tau_3$, in 6 different tasks. The attitude is represented in the ZYX-Euler angle for intuitive understanding, where roll (red) is the rotation around X-axis, pitch (green) is the rotation around Y-axis, yaw (blue) is the rotation around Z-axis. The overall attitude error is represented in the rotation error function $\Psi$, provided in \eqref{eq:ErrorFunc}. }
      \label{figure:AttitudeShow}
\end{figure*}

\begin{table}[t]
\caption{Attitude tracking parameters}
  \centering
  \label{tab:ATpara}
\begin{tabular}{@{}ll@{}}
\toprule
Parameters                                                               & Values                                               \\ \midrule
\multicolumn{2}{l}{\cellcolor[HTML]{EFEFEF}Comparison controller $\bm \tau_1$}                                                  \\ \midrule
attitude error coefficient, $G \in \mathbb R ^3$                         & ${\rm{diag}}\left( {\left[ {1~1~1} \right]^\top} \right)$ \\
attitude error gain, $k_R \in \mathbb R$                                 & 2                                                    \\
angular velocity error gain, $k_\Omega \in \mathbb R$                    & 0.2                                                  \\ \midrule
\multicolumn{2}{l}{\cellcolor[HTML]{EFEFEF}Comparison controller, $\bm \tau_2$}                                                 \\ \midrule
robust term $\bm v_1$ first coefficient, $\delta_b \in \mathbb R$     & 0.2                                                  \\
robust term $\bm v_1$ second coefficient, $\varepsilon \in \mathbb R$ & 0.1                                                  \\
adaptive term coefficient, $k_J \in \mathbb R$                             & 0.1                                                  \\
adaptive term damping, $\sigma \in \mathbb R$                             & 20                                                   \\ \midrule
\multicolumn{2}{l}{\cellcolor[HTML]{EFEFEF}Proposed controller, $\bm \tau_3$}                                                   \\ \midrule
robust term $\bm v_2$ gain, $k_v \in \mathbb R$                          & 0.15                                                 \\
robust term $\bm v_2$ exponent, $\rho_v \in \mathbb R$                     & 0.5                                                  \\ \bottomrule
\end{tabular}
\end{table}
The comparison simulations are conducted with 6 different desired attitude trajectories:
\begin{enumerate}
\item[ (a) ]  Start from Roll$=0~{\rm rad}$, Pitch$=0~{\rm rad}$, Yaw$=0~{\rm rad}$, with a constant desired angular velocity $[0.5 \pi ~ 0~ 0]^{\top}~{\rm rad/s}$.
\item[ (b1) ]  Start from Roll$=0~{\rm rad}$, Pitch$=0~{\rm rad}$, Yaw$=0~{\rm rad}$, with a constant desired angular velocity $[0~ 0.5 \pi~0]^{\top}~{\rm rad/s}$.
\item[ (b2) ] Start from Roll$=0~{\rm rad}$, Pitch$=0~{\rm rad}$, Yaw$=0~{\rm rad}$, with a constant desired angular velocity $[0~ -0.5 \pi~0]^{\top}~{\rm rad/s}$.
\item[ (c) ] Start from Roll$=0~{\rm rad}$, Pitch$=0~{\rm rad}$, Yaw$=0~{\rm rad}$, with a constant desired angular velocity $[0~ 0~0.5 \pi]^{\top}~{\rm rad/s}$.
\item[ (d) ] Start from Roll$=0~{\rm rad}$, Pitch$=0~{\rm rad}$, Yaw$=0~{\rm rad}$, although with a time-varying sinusoidal angular velocity $\left[ {\frac{\pi }{2}\cos (\pi t)~\frac{\pi }{2}\cos (\pi t )~\frac{\pi }{2}\cos (\pi t )} \right]^{\top}$.
\item[ (e) ] The desired attitude trajectory presents step changes to the next value after every $2000~{\rm ms}$, with trivial desired angular velocity constantly set as $[0~ 0~ 0]^{\top}~{\rm rad/s}$. The attitude values are given in ZYX-Euler angles, with the order as Roll, Pitch, Yaw: $\left[ {\frac{\pi }{2}~0~0} \right]^{\top}$, $\left[ {-\frac{\pi }{2}~0~0} \right]^{\top}$, $\left[0~{\frac{\pi }{2}~0} \right]^{\top}$, $\left[0~-{\frac{\pi }{2}~0} \right]^{\top}$,$\left[0~0~ -{\frac{\pi }{2}} \right]^{\top}$,$\left[0~0~ {\frac{\pi }{2}} \right]^{\top}$.
\end{enumerate}

The frequency of the controller calculation loop is 100Hz, one tenth of the simulation computation frequency, which is practical to implement in real flight experiments.
Simulation results are shown in Fig.~\ref{figure:AttitudeShow}. Parameters for these three controllers are all carefully tuned to 
achieve best performances, and are fixed for all 6 simulation tasks, which are clearly given in TABLE~\ref{tab:ATpara}. 

\begin{table}[t]
\caption{Attitude tracking steady state performance \\ measured in $\Psi$}
  \centering
  \label{tab:ATRes}
\begin{tabular}{ccccc}
\toprule
\multirow{2}{*}{Tasks}                                           & \multicolumn{1}{l}{\multirow{2}{*}{Indicators}} & \multicolumn{3}{c}{Control strategies}                                                                      \\ \cline{3-5} 
                                                                 & \multicolumn{1}{l}{}                            & \multicolumn{1}{l}{$\bm \tau_1$} & \multicolumn{1}{l}{$\bm \tau_2$} & \multicolumn{1}{l}{$\bm \tau_3$ Pro.} \\ \hline
\multirow{2}{*}{\ref{figure:AttitudeShow}-(a)}  & MAX                                             & 0.2315                           & 0.1821                           & \textbf{0.1060}                       \\
                                                                 & RMS                                             & 0.1043                           & 0.0775                           & \textbf{0.0475}                       \\
\multirow{2}{*}{\ref{figure:AttitudeShow}-(b1)} & MAX                                             & 0.1211                           & 0.1154                           & \textbf{0.0565}                       \\
                                                                 & RMS                                             & 0.0459                           & 0.0511                           & \textbf{0.0286}                       \\
\multirow{2}{*}{\ref{figure:AttitudeShow}-(b2)} & MAX                                             & \textbf{0.9031}                  & 1.0130                           & 1.3708                                \\
                                                                 & RMS                                             & \textbf{0.2808}                  & 0.3374                           & 0.4465                                \\
\multirow{2}{*}{\ref{figure:AttitudeShow}-(c)}  & MAX                                             & 0.1973                           & \textbf{0.0876}                  & 0.1100                                \\
                                                                 & RMS                                             & 0.0807                           & \textbf{0.0481}                  & 0.0550                                \\
\multirow{2}{*}{\ref{figure:AttitudeShow}-(d)}  & MAX                                             & 0.0866                           & 0.0842                           & \textbf{0.0510}                       \\
                                                                 & RMS                                             & 0.0358                           & 0.0333                           & \textbf{0.0221}                       \\
\multirow{2}{*}{\ref{figure:AttitudeShow}-(e)}                     & MAX                                             & 0.2171                           & 1.4197                           & \textbf{0.1140}                       \\
                                                                 & RMS                                             & 0.0509                           & 0.4208                           & \textbf{0.0236}                       \\ \bottomrule
\end{tabular}
\end{table}
\unskip
Generally, the control strategy $\bm \tau_3$ superiors over the other two, concluding from different tasks attitude errors shown in Fig.~\ref{figure:AttitudeShow}.
The control strategy $\bm \tau_1$ can barely satisfactorily achieve the attitude tracking task, which is however easy-to-use owing to 
its fewer parameters. The control strategy $\bm \tau_2$ can achieve satisfactory performance in steady flight although still having larger attitude error compared with $\bm \tau_3$. Due to the integral essence of the controller $\bm \tau_2$ adaptive terms, implementing it in a flapping wing robot, which is oscillating and has no constant inertia matrix, system error accumulation may consequently induce unexpected behaviors, for example, the two failures shown in Fig. \ref{figure:AttitudeShow}-(e). Based on these observations, $\bm \tau_3$ is selected as the attitude tracking control strategy for following tasks and applications. The steady-state performances are shown in TABLE \ref{tab:ATRes}, which also indicate that controller $\bm \tau_3$ lead to better attitude trajectory tracking performance.
In this table, MAX indicates the maximum attitude error $\psi$, RMS indicates the root mean square of $\psi$. 

The following two behaviors in the attitude tracking tasks are worthy to be deeply investigated.
\begin{figure}[t]
      \centering
      \includegraphics[width=2.7in]{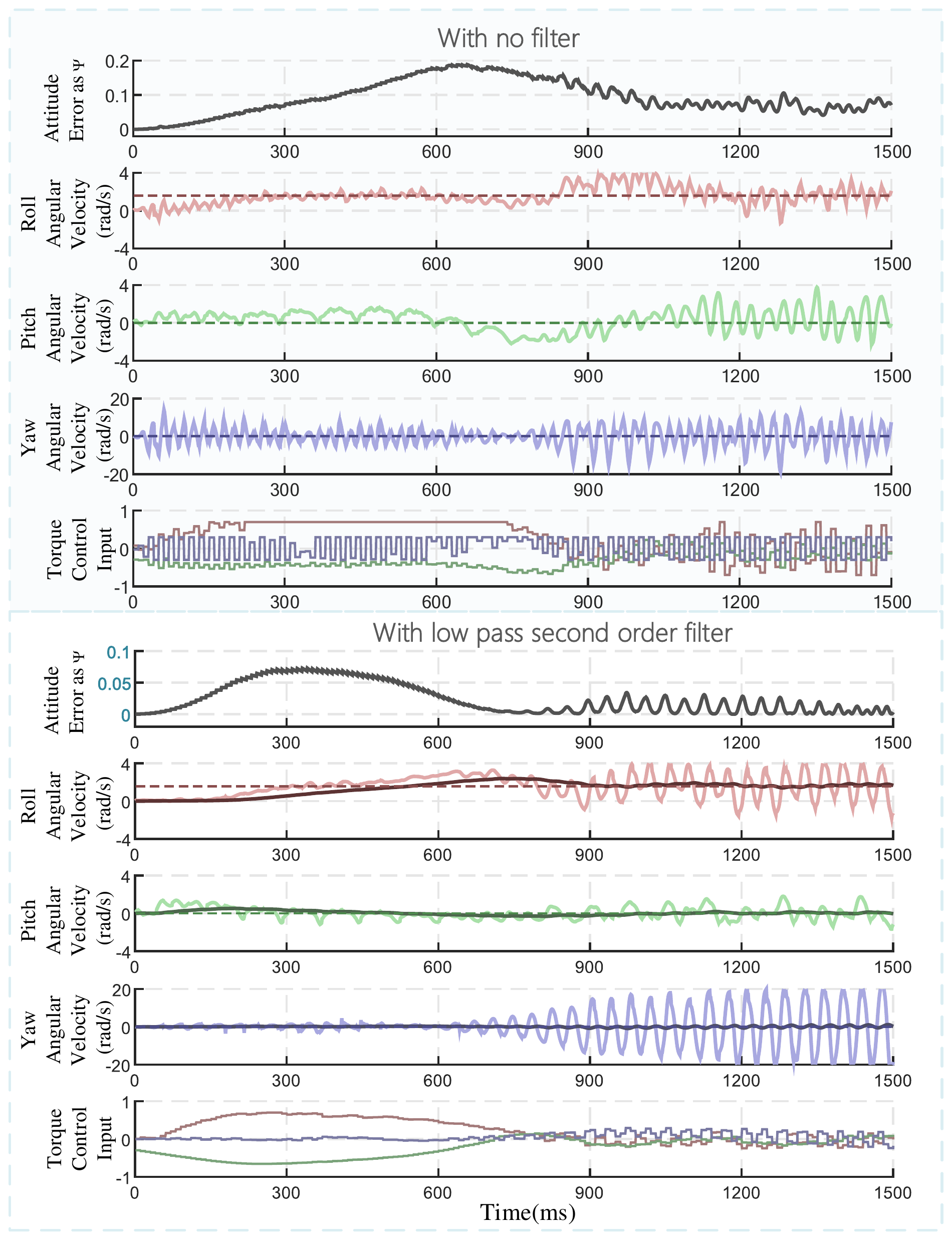}\\
      \caption{Attitude tracking control performance without and with filter following desired attitude trajectory (a) equipped with control strategy $\bm \tau_3$. The darker color curves represent the filtered signals, while the lighter curves are the raw signals.}
      \label{figure:FilterConShow}
\end{figure}
\unskip
\begin{figure}[t]
      \centering
      \includegraphics[width=2.7in]{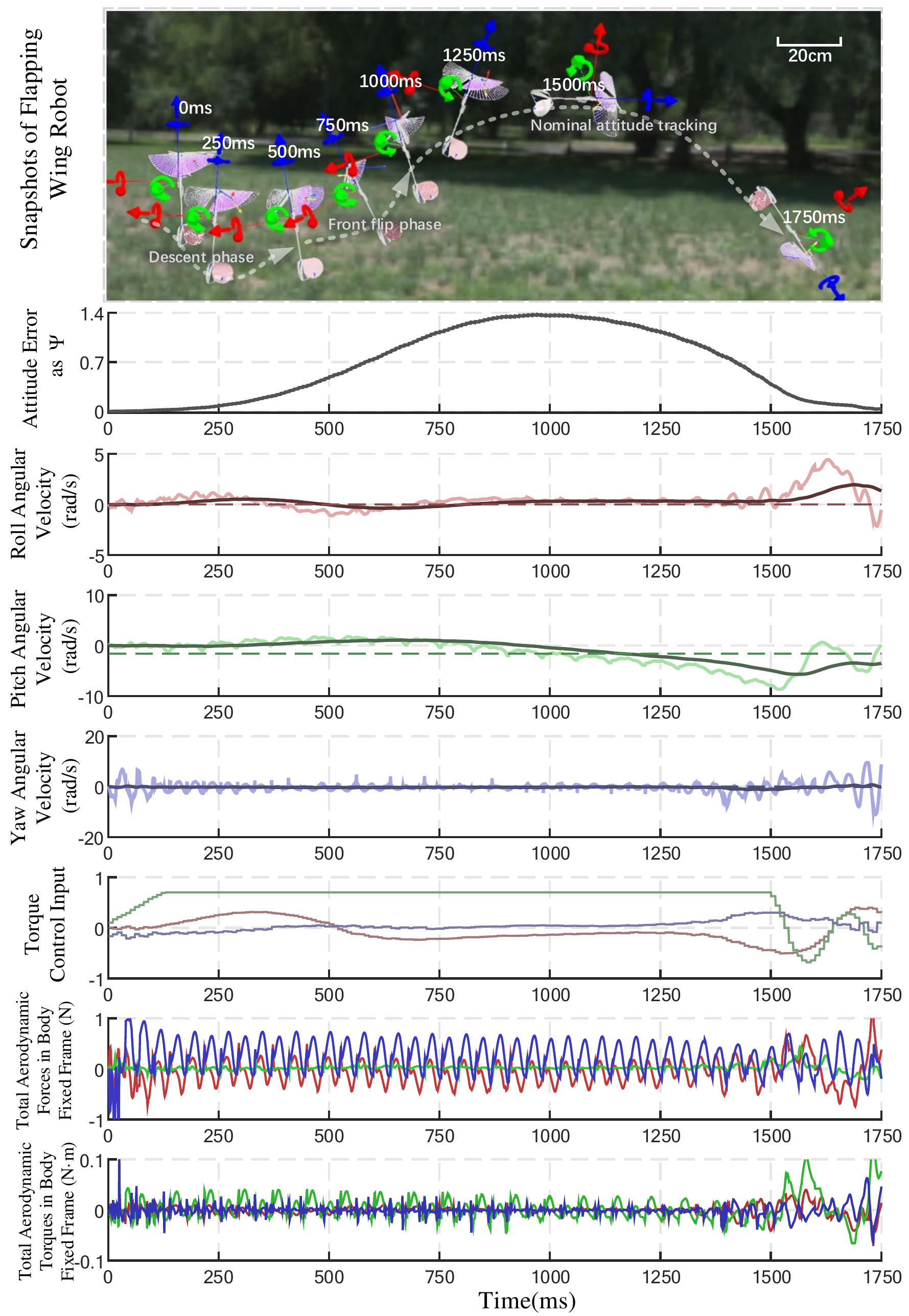}\\
      \caption{Flapping wing robot behavior equipped with control strategy $\bm \tau_3$, when tracking desired attitude trajectory (b2) where the unexpected large attitude tracking error emerges.}
      \label{figure:UnexpectShow}
\end{figure}
\unskip
\begin{enumerate}
  \item[ $\bullet$ ] BAT-1: There are obvious angular velocity oscillations in almost all the tasks, especially for the yaw kinematics. 
                            Observing the variation trend of the angular velocities in Fig. \ref{figure:AttitudeShow}-(a)(b1)(b2), it
                            can be concluded that, when the robot is upright with relatively small free flow, the oscillation is moderate,
                            and becomes drastic when the surrounded free flow is prominent, especially when the vehicle is faced with the circumstance as Fig. \ref{figure:AveShowVel}-(a1)(a2)(b1)(b2).
                            As shown in the comparison simulation shown in Fig. \ref{figure:FilterConShow}, where the robot follows the attitude 
                            trajectory (a) equipped with controller $\bm \tau_3$, the robot can achieve better performances with designed filter,
                            with an error reduction larger than $60 \%$.
                            Moreover, the control inputs drastically oscillate without filtered feedback signals, which leads to larger control efforts, while the filtered ones are relatively steady, which is approximately $50 \%$ of the oscillation amplitude without the filter.
                            In summary, the controller with 100Hz computation frequency, as well as the practical actuators, \emph{e.g.} servo motor usually only capable of achieving ${\pi  \mathord{\left/{\vphantom {\pi  {\rm{3}}}} \right.\kern-\nulldelimiterspace} {\rm{3}}}$ steering in at least $0.1~{\rm s}$,
                            cannot actually be used to suppress the 13Hz frequency flapping induced oscillation. To this end, the oscillating angular velocity signal is hazardous in the feedback loop, which can drown the real effective feedback signal. 
                            On the other hand, after implementing the filter, the system being stable in the average sense, tolerable oscillation is certainly not taken into the feedback loop, therefore has no chance and is not necessary to be suppressed by the controller.
  \item[ $\bullet$ ] BAT-2: The unexpected large errors occur in the simulation of all three controllers, when the vehicle tracks trajectory (b2),
                            where the robot is basically back-flipping. Generally, the behavior difference between front-flip shown in Fig.~\ref{figure:AttitudeShow}-(b1) and back-flip shown in Fig. \ref{figure:AttitudeShow}-(b2) is mainly due to PW-2.
                            Let us further investigate the unexpected behaviors in $0$-$500~{\rm ms}$ and $500$-$1000~{\rm ms}$, shown in Fig.~\ref{figure:UnexpectShow}, what we call \emph{Descent  phase} and \emph{Front flip phase}, respectively.
                            In \emph{Descent phase}, the control input signal shown in Fig.~\ref{figure:UnexpectShow} indicates that the robot 
                            has a similar situation demonstrated in Fig.~\ref{figure:AveShow}-(h), where the robot suffers from PW-1. 
                            At the same time, the rotated thrust force makes the robot accelerate backward. After the acceleration, and reaching a backward velocity approximately to $1~{\rm m/s}$, the robot comes into \emph{Front flip phase}. 
                            In this phase, PF-2 dominates, such that the robot pitches forward due to the airflow from the dorsal side.
                            Combined with PW-2, the robot pitch is relatively fast. 
                            This pitch rotates the thrust back toward the opposite direction of gravity, making the robot ascend. 
                            In the meantime, the backward thrust component reduces, and due to the resistance, the backward velocity decreases, thus PF-2 vanishes. Then the robot can finally generate negative pitch torque to track its desired attitude trajectory, and \emph{Front flip phase} therefore ends.

\end{enumerate}

\subsection{Trajectory Tracking}
Trajectory tracking in 3-dimensional Euclidean space is essential to explore the vehicle dynamics feasibility in
the acrobatic flight, as well as the vehicle robustness in conventional flight.
The control objective is to make the robot position following the desired translational trajectory, which is 
\begin{equation}
\label{eq:DesireP}
{\dot {\bm p}_d} = {\bm v} _d
\end{equation}
where ${ \bm  v_d} \in \mathbb R^3$ is the desired translational velocity in the inertia frame, 
${ \bm  p_d} \in \mathbb R^3$ is the desired 3-dimensional position. 

The translational dynamics can be given as 
\begin{align}
\label{eq:Tra-dynamics}
m\dot {\bm v}& = R \cdot F_{t} \bm e_3 + R{\bm F_{a}}\left( \bm Z_b \right) + \delta {\bm F} - mg{\bm e_3},\\
\dot {\bm p} &= {\bm v}\nonumber
\end{align}
where $m \in \mathbb {R}$ is the robot mass, ${\bm p}, {\bm v} \in \mathbb {R}^3$ are the translational position and velocity of the robot mass center represented in the inertia frame, respectively, ${\bm e_3} = [0~~0~~1]^{\top} \in \mathbb {R}^3$ represents the vertical direction in the inertia frame,
$F_{t} \in \mathbb {R}$ is the aerodynamics force component along the cephalad direction generated by the flapping wings, 
$\bm F_{a} (\bm Z_b) \in \mathbb {R}^3$ is the remainder aerodynamics forces, ${\bm Z_b} = [ F~~{\bm v}^{\top}~~{\bm \Omega} ^{\top} ]^{\top} \in \mathbb R ^ 7$ is the composition of aerodynamics influence factors, all resolved in the inertia frame, $\delta {\bm F} \in \mathbb {R}^3$ is the composition of the force disturbances and actuator faults from model uncertainties and actuator misalignment,
$g \in \mathbb {R}$ represents the gravitational acceleration magnitude. 
According to the robot attitude dynamics, the virtual control inputs of the translational dynamics system are the rotation matrix $R$ and the thrust $F$.

The translation error function is defined as
\begin{equation}
\label{eq:Tra-errors}
{\bm e_p}\left( t \right) = {\bm p_d}\left( t \right) - \bm p\left( t \right)
\end{equation}
where ${\bm e_p} \in \mathbb {R}^3$ is the robot translational error.

The translational error dynamics is shown as 
\begin{align}
&{{\dot {\bm e}}_p} = { {\bm v}_d} - {\bm v},\nonumber\\
&{{\bm e}_v}: = {{\dot {\bm e}}_p}, \nonumber\\
&{{\dot {\bm e}}_v} = {{\dot {\bm v}}_d} - \frac{1}{m}R ({F_t}{{\bm e}_3}) - \frac{1}{m}R{{\bm F}_a}\left( {{{\bm Z}_b}} \right)-\frac{1}{m}\delta {\bm F} + g{{\bm e}_3}
\label{eq: error-dynamics}
\end{align}
where ${\bm e_v} \in \mathbb {R}^3$ is the translational velocity error.

\begin{Remark}
\label{Re:flapp_pos_pro}
The position trajectory tracking control problem of the flapping wing robot has the following three challenges: 
\begin{enumerate}
\item  Available control inputs appearing as $\frac{1}{m}R \cdot {F_t}{{\bm e}_3}$ only possess comparable magnitude of 
$\frac{1}{m}R{{\bm F}_a}\left( {{{\bm Z}_b}} \right)$. And these two terms are coupled, indicating that 
when the robot attitude as the control input changes, the remainder aerodynamic forces will also change. 
Therefore, we have to utilize $\frac{1}{m}R{{\bm F}_a}\left( {{{\bm Z}_b}} \right)$ to achieve our control objective, instead of rejecting it.
Since $\frac{1}{m}R{{\bm F}_a}\left( {{{\bm Z}_b}} \right)$ is nearly not possible to be accurately modeled, the imperative need is to efficiently estimate it online.

\item  How to reject the undesired $\delta \bm F$ disturbances as much as possible, by practical inputs and actuators.

\item  Recalling BAT-2, when the robot translational velocity is deviated from the robot cephalad direction and is simultaneously large in magnitude, the unexpected behavior will emerge.
Therefore, the robot cephalad direction should be constrained within a time-varying domain depending on translational velocity direction resolved in the body-fixed frame, in order to maintain maneuverability.
\end{enumerate}
\end{Remark}


Let us simplify the expression of \eqref{eq:Tra-dynamics} and \eqref{eq: error-dynamics}. Taking $\frac{1}{m}R ({F_t}{{\bm e}_3})$ as $\bm u_t \in \mathbb R_3$, $\frac{1}{m}R{{\bm F}_a}\left( {{{\bm Z}_b}} \right)$ as $\bm f_a(t) \in \mathbb R_3$, the positional dynamics can be  rewritten as 
\begin{align}
\label{eq:Tra-dynamics_sim}
m\dot {\bm v}& = {\bm u_t}+ m{\bm f_a}\left( t \right) + \delta {\bm F} - mg{\bm e_3},\\
\nonumber
{{\dot {\bm e}}_v} &= {{\dot {\bm v}}_d} - {\bm f_a}\left( t \right) - \frac{1}{m}\delta \bm F + g{\bm e_3} - {\bm u_t}
\end{align}

Based on \eqref{eq:Tra-dynamics_sim} and enlightened by \cite{Zhao-2013}, an extended state observer is used to attack the first challenge in \emph{Remark~\ref{Re:flapp_pos_pro}}. Since only ${\bm p }$ is observable, the observer is designed as
\begin{align}
\label{eq:Observer_1}
\dot { {\bm p}}_{o} &=  {\bm v}_{o} - {G_p}{\bm \sigma ^{\frac{{{\rho_e} + 1}}{2}}}\left( { {\bm p}_{o} - {\bm p}} \right)\\
\label{eq:Observer_2}
m\dot{ {\bm v}}_{o} &= {\bm u_t} - mg{\bm e_3} - {G_v}{\bm\sigma ^{\frac{{{\rho_e} + 1}}{2}}}\left( { {\bm p}_{o} - {\bm p}} \right) +  {\bm z}\\
\label{eq:Observer_3}
\dot { {\bm z}} &=  - {G_{z}}{\bm\sigma ^{{\rho_e}}}\left( {{\bm p}_{o} - {\bm p}} \right)
\end{align}
where $ {\bm z} \in \mathbb{R}^3$ can be viewed as an estimation of $m{\bm f_a}\left( t \right) + \delta {\bm F}$, 
$ {\bm p}_{o} \in \mathbb{R}^3$ and ${\bm v}_{o} \in \mathbb{R}^3$ are the estimations of ${\bm p} $ and ${\bm v}$, respectively,
$G_p, G_v, G_z \in \mathbb{R}^{3 \times 3}$ are positive definite, diagonal matrices, and $\bm\sigma^{\star_1}(\bm \star_2): \mathbb{R} \times \mathbb{R}^3 \to \mathbb{R}^3$ is a 
vector function, which implements the operation ${\rm sgn}(\star_{2i}) \left| \star_{2i}\right|^{\star_1}$ on each entry $\star_{2i}$ of vector $\bm \star_2$, then sequentially connect obtained results as the output vector, $0 < \rho_e < 1\in \mathbb R$ is a positive constant.
It is noteworthy that, the robot velocity is not directly available, such that its estimation ${\bm v}_o$ is 
used as the feedback.

Then we use the sliding mode robust control technique to overcome the second challenge in \emph{Remark~\ref{Re:flapp_pos_pro}}. 
The control law is developed as 
\begin{align}
\label{eq:SlidContorl}
{\bm u_{t1}} =& \underbrace {{K_s}{\bm \sigma ^{{\rho _s}}}\left(\bm  s \right) + {K_{ep}}{ {\mathbf {tanh}}}({K_p { {\bm e}}_p}) + {K_{ev}}{{ {\bm e}}_v} + {K_{eI}}R{{ {\bm e}}_I}}_{{\rm{Feedback~terms}}} \nonumber\\
& + \underbrace {{{\dot {\bm v}}_d} - \frac{1}{m} {\bm z} + g{\bm e_3}}_{{\rm{Feedforward~terms}}}
\end{align}
where $\bm s = c_s{ {\mathbf {tanh}}}({K_p { {\bm e}}_p}) + {K_v}{{ {\bm e}}_v} \in \mathbb{R}^3$, ${{ {\bm e}}_I} \in \mathbb R^3$ is the integral feedback considering both velocity and positional errors in the body-fixed frame, updated by ${\dot {\bm e}_I} = {\bf{proj}}_{{e_{{{Ib}}}}}\left( R^\top ({{K_{Ip}}{{\bm e}_p} + {K_{Iv}}{{\bm e}_v}}) \right)$ and ${{ {\bm e}}_p} = {\bm p_d} -  {\bm p}_{o}$, ${{ {\bm e}}_v} = {\bm v_d} -  {\bm v}_{o}\in \mathbb{R}^3$ are the position and velocity errors computed by the filtered signal, respectively, 
${\bf{proj}}_{{e_{{{Ib}}}}}\left( \bm \star\right): \mathbb R^3 \to \mathbb R^3$ is the projection function which 
constrains $\bm e_{I}$ in predefined bound $e_{Ib}$.
and $K_s, K_p,  K_v, K_{ep}, K_{ev}, {K_{eI}}, {K_{Ip}, {K_{Iv}}}\in \mathbb{R}^{3 \times 3}$ are positive definite, diagonal matrices, $c_s \in \mathbb R$ is a positive constant. The saturation function $ {\mathbf {tanh}}({\bm \star}):  \mathbb{R}^3 \to \mathbb{R}^3$ is a 
vector function, which implements the hyperbolic tangent function operation ${\rm tanh}(\star) $ on each entry $\star$ of vector $\bm \star$, and sequentially connects their results, moreover, $0 < \rho_s < 1\in \mathbb R$ is a positive constant. 
The saturation function is implemented to avoid large position errors overwhelming other signals.
\subsection{Desired Attitude Trajectory Generation}
The nominal virtual input $\bm u_t$ can be composed of the thrust magnitude and orientation.  
Based on Rodrigues' rotation formula, the desired orientation without rotation around Z-axis in the body-fixed frame, ${R_{d\backslash Z }}\in SO(3)$, can be 
developed as 
\begin{align}
\label{eq:Rodrigues}
&{R_{d\backslash Z }} = I + \hat {\bm k} + \frac{{1 - c}}{{{s^2} + \varepsilon_{k}}}{{\hat {\bm k}}^2},\\
&{\bm k}= {{\bm e}_3} \times {{\bar {\bm u}}_t},~c = {{\bm e}_3} \cdot {{\bar {\bm u}}_t},~s = \left\| {\bm k} \right\|,~{{\bar {\bm u}}_t} = {{{\bm u}_t}}/{{\left\| {{{\bm u}_t}} \right\|}}\nonumber
\end{align}
where ${\bm k} \in \mathbb R^3$ can be viewed as the rotation axis multiplied with sine value $s\in \mathbb R$ of the rotation angle,
$\hat {\bm k} \in \mathbb R^{3 \times 3}$ is the skew symmetric matrix transformed from ${\bm k}$,
$\varepsilon_{k} \in \mathbb R ^+$ is a small positive constant to avoid singularity,
${{\bar {\bm u}}_t} \in \mathbb R^3$ is the normalized ${\bm u}_t$, $c\in \mathbb R$ is the cosine value of the rotation angle.
To a certain extent, the rotation around Z-axis in the body-fixed frame is free in composing $\bm u_t$, whose rotation matrix conformed to the following restriction 
\begin{equation}
{R_Z}\left( \alpha  \right) = \left[ {\begin{array}{*{20}{c}}
{\cos \alpha }&{ - \sin \alpha }&0\\
{\sin \alpha }&{\cos \alpha }&0\\
0&0&1
\end{array}} \right]
\nonumber
\end{equation}
where $\alpha \in \mathbb R$ is the rotation angle.
Then the desired orientation as a rotation matrix is given by 
\begin{equation}
R_d = {R_{d\backslash Z}}{R_Z}
\end{equation}
Recalling that there exists velocity term $\bm \Omega_d$ in the error dynamics \eqref{eq:ErrorOmega}, 
the continuity of ${R_{d\backslash Z}}{R_Z}$ above the second differential is indispensable,
which is addressed by alternatively using the signal from a differential tracker
\begin{align}
{{\dot R}_f} &= {R_f}{{\hat {\bm \Omega} }_f}\nonumber\\
{{\dot {\bm \Omega} }_f} &=  - {K_{\omega f}}{{\bm \Omega} _f} - {k_{Rf}}{{\bm e}_{Rf}}\nonumber\\
{{\bm e}_{Rf}} &= \frac{1}{2}{\left( {{G_f}R_d^\top{R_f} - R_f^\top{R_d}G_f} \right)^ \vee }
\label{eq:Velocity_Tracker}
\end{align}
where $k_{Rf} \in \mathbb R$ is positive constant, and ${K_{\omega f}}, G_f \in \mathbb R^3$ are positive definite, diagonal 
matrices, $R_f \in SO(3)$ is the filtered rotation matrix, $\bm \Omega_f \in \mathbb R^{3 \times 3}$ is the filtered angular velocity,
and ${{\bm e}_{Rf}} \in \mathbb R^3$ is the attitude error between the filtered and desire.
Furthermore, the thrust is given by $m\left\| {{\bm u_t}} \right\|$.
And we can use the robot hovering flight flapping wing frequency to translate the thrust into flapping frequency, considering that the flapping frequency is linear with respect to the thrust, where the modeling error can be handled by
the robust controller:
\begin{equation}
f_t\left( t \right) = \frac{{{f_{{\rm{hover}}}}}}{g}\left\| {{\bm u_t}\left( t \right)} \right\| \nonumber
\end{equation}
where $f_{\rm{hover}} \in \mathbb R^+$ is the flapping frequency at hovering. In order to maintain the pitch and yaw torque 
generation capability, the flapping wing frequency is restricted in $[9,15]$~Hz.

Since the robot is susceptible to wind direction with limited attitude maneuverability, as illustrated in PF-2 and PF-3, the planning of the $R_Z$, in comparison to quadrotors and symmetric flapping wing robots, suffers more restrictions.
Similar to the design of $R_{d\backslash Z}$, we can construct the desire rotation $R_Z$ in the following way:
\begin{align}
\label{eq:Rodrigues2}
&{R_{Z}} = I + \hat {\bm k}_z + \frac{{1 - c_z}}{{{s^2_z}}}{{\hat {\bm k}}^2_z},\\
&{\bm k_z}= {{\bm d}_T} \times {{\bar {\bm v}}_{xy}},~c_z = {{\bm d}_T} \cdot {{\bar {\bm v}}_{xy}},~s_z = \left\| {\bm k} \right\|,~{{\bar {\bm v}}_{xy}} = {{{\bm v}_{xy}}}/{{\left\| {{{\bm v}_{xy}}} \right\|}}\nonumber
\end{align}
where ${\bm v}_{xy} = {\left[ {{v_{dx}}{\rm{~}}{v_{dy}}{\rm{~}}{0}} \right]^\top} \in \mathbb R^3$ is the desired velocity direction in inertia frame XY-plane, while ${\bm v_d} = {\left[ {{v_{dx}}{\rm{~}}{v_{dy}}{\rm{~}}{v_{dz}}} \right]^\top}$, ${\bm d}_T \in \mathbb R ^3$ is the direction aligned to the desired velocity direction in the XY-plane, which can be designed to determine the flight mode.

\begin{Remark}
\label{Re:high_freq_filter}
Expecting to perform aggressive flights, it is noteworthy that the observer (\ref{eq:Observer_1}-\ref{eq:Observer_3}) and the filter \eqref{eq:Velocity_Tracker} are necessary to be updated in a higher frequency than the controller. Furthermore, 
the filter \eqref{eq:Velocity_Tracker} is actually running discretely, thus normalization of the rotation matrix is needed, whose accuracy is also dependent on high-frequency update. Therefore, the observer and the filter are updated at 1000Hz.
\end{Remark}

\begin{figure}[t]
      \centering
      \includegraphics[width=3.2in]{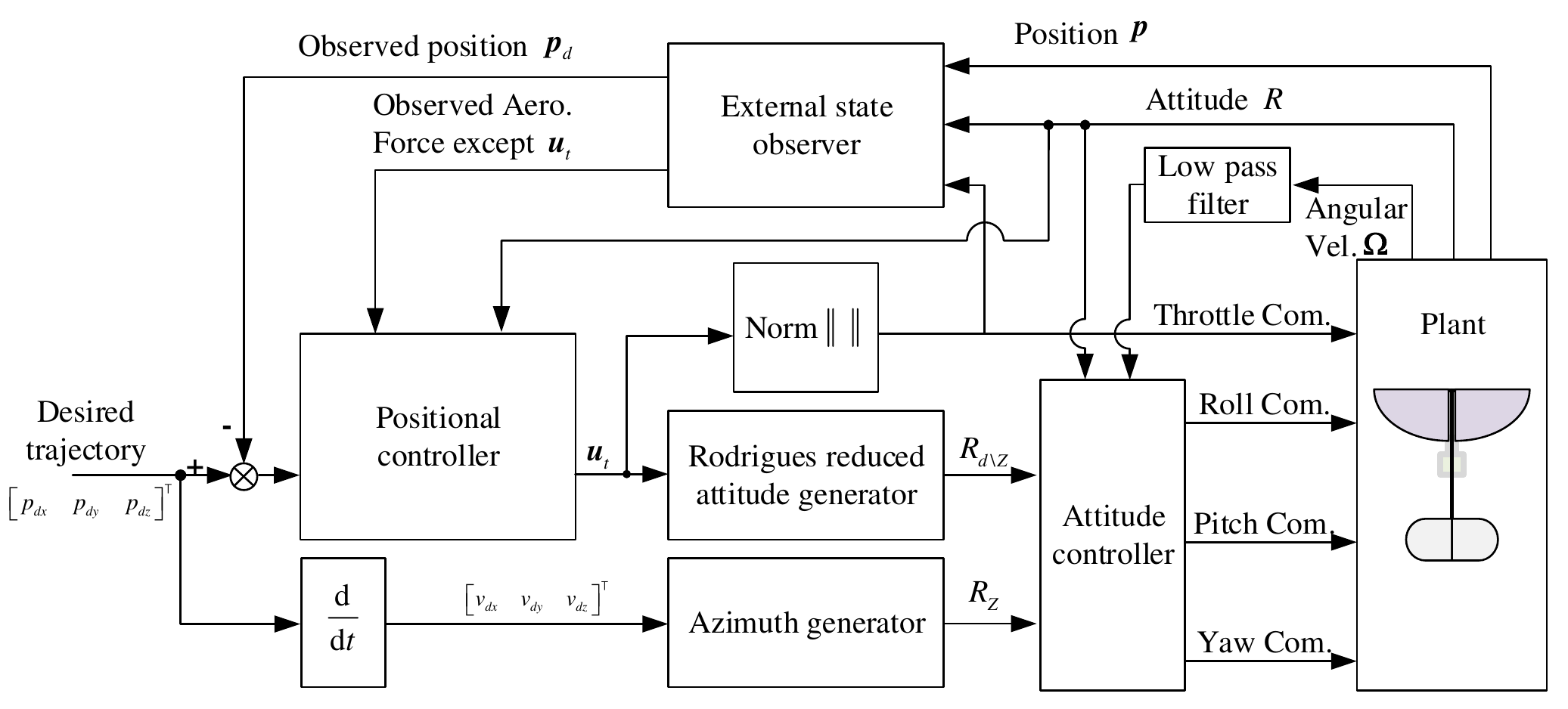}\\
      \caption{Flapping wing robot trajectory tracking control block diagram.}
      \label{figure:Boxdia}
\end{figure}

In conclusion, the position controller, attitude controller, external state observer, and filter are synthesized into the 
trajectory controller, whose overall diagram is shown in Fig.~\ref{figure:Boxdia}

\subsection{Autonomous Flight Simulation}
The flapping wing robot autonomous flight with the proposed control scheme is conducted in our simulation platform.
The attitude tracking controller ${\bm \tau _3}$ shown in \eqref{eq:pro_at_con} is used in the simulation. 
For comparison, the following trajectory tracking controller are also implemented.
\begin{equation}
\label{eq:NonlinearPid}
\bm u_{t2} = K_{ep2}{\bf {tanh}} \left( K_{p2}{\bm e_p} \right) + K_{ev2}{\bm e_v} 
\end{equation}
\begin{equation}
\label{eq:Slidemode}
{\bm u_{t3}} = {K_{s3}}{\bf {  sgn}} \left( {{\bm s_3}} \right) + {K_{ep3}}{\bm e_p} + {K_{ev3}}{\bm e_v} + {{\dot {\bm v}}_d} + g{\bm e_3}
\end{equation}
where ${\bm s_3} = {\bm e_p} + {K_{v3}}{\bm e_v}$, and ${K_{ep2}}$, ${K_{p2}}$, ${K_{ev2}}$, ${K_{s3}}$, ${K_{ep3}}$, ${K_{ev3}}, {K_{v3}} \in \mathbb R_{3 \times 3}$ are positive definite, diagonal matrices. Generally, the control law ${\bm u_{t2}}$
can be classified as the model reference saturated PD controller, which is similar to the position controller proposed in \cite{He-2021}, and the control law ${\bm u_{t3}}$ can be seen as a sliding mode controller, which is analogue to the controller proposed in \cite{Bluman-2018}. 

\begin{table}[t]
\caption{Trajectory tracking parameters}
  \centering
  \label{tab:TTpara}
\begin{tabular}{@{}ll@{}}
\toprule
Parameters                                                                                                                           & Values                                                         \\ \midrule
\multicolumn{2}{l}{\cellcolor[HTML]{EFEFEF}Proposed controller $\bm u_{t1}$}                                                                                                                           \\ \midrule
robust term gain, $K_s \in \mathbb R ^{3 \times 3}$                                                                                  & ${\rm{diag}}\left( {\left[ {1~1~1} \right]^\top} \right)$      \\
\begin{tabular}[c]{@{}l@{}}positional error gain in robust term \\ ~before saturation, $K_p \in \mathbb R ^{3 \times 3}$\end{tabular} & $0.8~{\rm{diag}}\left( {\left[ {1~1~1} \right]^\top} \right)$  \\
\begin{tabular}[c]{@{}l@{}}velocity error gain \\ ~in robust term, $K_v \in \mathbb R ^{3 \times 3}$\end{tabular}                     & $0.5~{\rm{diag}}\left( {\left[ {1~1~1} \right]^\top} \right)$  \\
\begin{tabular}[c]{@{}l@{}}positional error gain \\ ~before saturation, $K_{ep} \in \mathbb R ^{3 \times 3}$\end{tabular}             & $10~{\rm{diag}}\left( {\left[ {1~1~1} \right]^\top} \right)$   \\
\begin{tabular}[c]{@{}l@{}}velocity error gain , \\ ~$K_{ev} \in \mathbb R ^{3 \times 3}$\end{tabular}                                & $0.5~{\rm{diag}}\left( {\left[ {1~1~1} \right]^\top} \right)$  \\
\begin{tabular}[c]{@{}l@{}}integral error gain, \\ ~$K_{eI} \in \mathbb R ^{3 \times 3}$\end{tabular}                                 & $0.01~{\rm{diag}}\left( {\left[ {1~1~1} \right]^\top} \right)$ \\
\begin{tabular}[c]{@{}l@{}}positional error gain \\ ~in integral term, $K_{Ip} \in \mathbb R ^{3 \times 3}$\end{tabular}              & $0.8~{\rm{diag}}\left( {\left[ {1~1~1} \right]^\top} \right)$  \\
\begin{tabular}[c]{@{}l@{}}velocity error gain in integral term,\\  ~$K_{Iv} \in \mathbb R ^{3 \times 3}$\end{tabular}                & $0.5~{\rm{diag}}\left( {\left[ {1~1~1} \right]^\top} \right)$  \\
\begin{tabular}[c]{@{}l@{}}positional error gain in robust term \\ ~after saturation, $c_s \in \mathbb R$\end{tabular}                & 2                                                              \\
robust term exponent, $\rho_s \in \mathbb R$                                                                                         & 0.5                                                            \\ \midrule
\multicolumn{2}{l}{\cellcolor[HTML]{EFEFEF}ESO used in $\bm u_{t1}$}                                                                                                                                  \\ \midrule
\begin{tabular}[c]{@{}l@{}}first order observer gain, \\ ~$G_p \in \mathbb R ^{3 \times 3}$\end{tabular}                              & $20~{\rm{diag}}\left( {\left[ {1~1~1} \right]^\top} \right)$    \\
\begin{tabular}[c]{@{}l@{}}second order observer gain, \\ ~$G_v \in \mathbb R ^ {3 \times 3}$\end{tabular}                            & $10~{\rm{diag}}\left( {\left[ {1~1~1} \right]^\top} \right)$   \\
\begin{tabular}[c]{@{}l@{}}extended state observer gain, \\ ~$G_z \in \mathbb R ^ {3 \times 3}$\end{tabular}                          & $5~{\rm{diag}}\left( {\left[ {1~1~1} \right]^\top} \right)$   \\
ESO exponent, $\rho_e \in \mathbb R$  & 0.5                                                            \\ \midrule
\multicolumn{2}{l}{\cellcolor[HTML]{EFEFEF}Comparison controller $\bm u_{t2} $}                                                                                                                       \\ \midrule
\begin{tabular}[c]{@{}l@{}}positional error gain after saturation, \\ ~$K_{ep2} \in \mathbb R ^{3 \times 3}$\end{tabular}             & $10~{\rm{diag}}\left( {\left[ {1~1~1} \right]^\top} \right)$   \\
\begin{tabular}[c]{@{}l@{}}positional error gain before saturation, \\ ~$K_{p2} \in  \mathbb R ^{3 \times 3}$\end{tabular}            & $0.8~{\rm{diag}}\left( {\left[ {1~1~1} \right]^\top} \right)$  \\
\begin{tabular}[c]{@{}l@{}}velocity error gain, \\ ~$K_{ev2} \in \mathbb R ^{3 \times 3}$\end{tabular}                                & ${\rm{diag}}\left( {\left[ {1~1~1} \right]^\top} \right)$      \\ \midrule
\multicolumn{2}{l}{\cellcolor[HTML]{EFEFEF}Comparison controller $\bm u_{t3}$}                                                                                                                        \\ \midrule
\begin{tabular}[c]{@{}l@{}}sliding mode robust term gain, \\ ~$K_{s3} \in \mathbb R ^{3 \times 3}$\end{tabular}                       & $0.5~{\rm{diag}}\left( {\left[ {1~1~1} \right]^\top} \right)$  \\
\begin{tabular}[c]{@{}l@{}}velocity error gain in robust term, \\ ~$K_{v3} \in \mathbb R ^{3 \times 3}$\end{tabular}                  & $0.5~{\rm{diag}}\left( {\left[ {1~1~1} \right]^\top} \right)$  \\
\begin{tabular}[c]{@{}l@{}}positional error gain, \\ ~$K_{ep3} \in \mathbb R ^{3 \times 3}$\end{tabular}                              & $8~{\rm{diag}}\left( {\left[ {1~1~1} \right]^\top} \right)$    \\
\begin{tabular}[c]{@{}l@{}}velocity error gain, \\ ~$K_{ev3} \in \mathbb R ^{3 \times 3}$\end{tabular}                                & $0.5~{\rm{diag}}\left( {\left[ {1~1~1} \right]^\top} \right)$  \\ \bottomrule
\end{tabular}
\end{table}

\begin{table}[t]
\caption{Trajectory tracking steady state performance\\ measured in positional error}
  \centering
  \label{tab:TTRes}
\begin{tabular}{lcccc}
\toprule
\multirow{2}{*}{Tasks}      & \multicolumn{1}{l}{\multirow{2}{*}{Indicators}} & \multicolumn{3}{c}{Control strategies}                                                                      \\ \cline{3-5} 
                            & \multicolumn{1}{l}{}                            & \multicolumn{1}{l}{$\bm u_{t1}$ Pro.} & \multicolumn{1}{l}{$\bm u_{t2}$} & \multicolumn{1}{l}{$\bm u_{t3}$} \\ \midrule
\multirow{2}{*}{(a) v-Cir.} & MAX                                             & \textbf{0.2314~m}                & 1.0235~m                    & 0.8512~m                    \\
                            & RMS                                             & \textbf{0.1226~m}                & 0.9400~m                    & 0.7681~m                    \\
\multirow{2}{*}{(b) l-Cir.} & MAX                                             & \textbf{0.3436~m}                & 0.7466~m                    & 2.0142~m                    \\
                            & RMS                                             & \textbf{0.1660~m}                & 0.6722~m                    & 0.8515~m                    \\
\multirow{2}{*}{(c) Lem.}   & MAX                                             & 0.8334~m                         & 0.9645~m                    & \textbf{0.8016~m}           \\
                            & RMS                                             & \textbf{0.4433~m}                & 0.8346~m                    & 0.6789~m                    \\
\multirow{2}{*}{(d) P2P}    & MAX                                             & \textbf{0.9644~m}                & 2.4993~m                    & 1.2115~m                    \\
                            & RMS                                             & \textbf{0.3499~m}                & 0.7726~m                    & 0.4674~m    
 \\ \bottomrule              
\end{tabular}
\end{table}

\begin{figure}[t]
      \centering
      \includegraphics[width=3.4in]{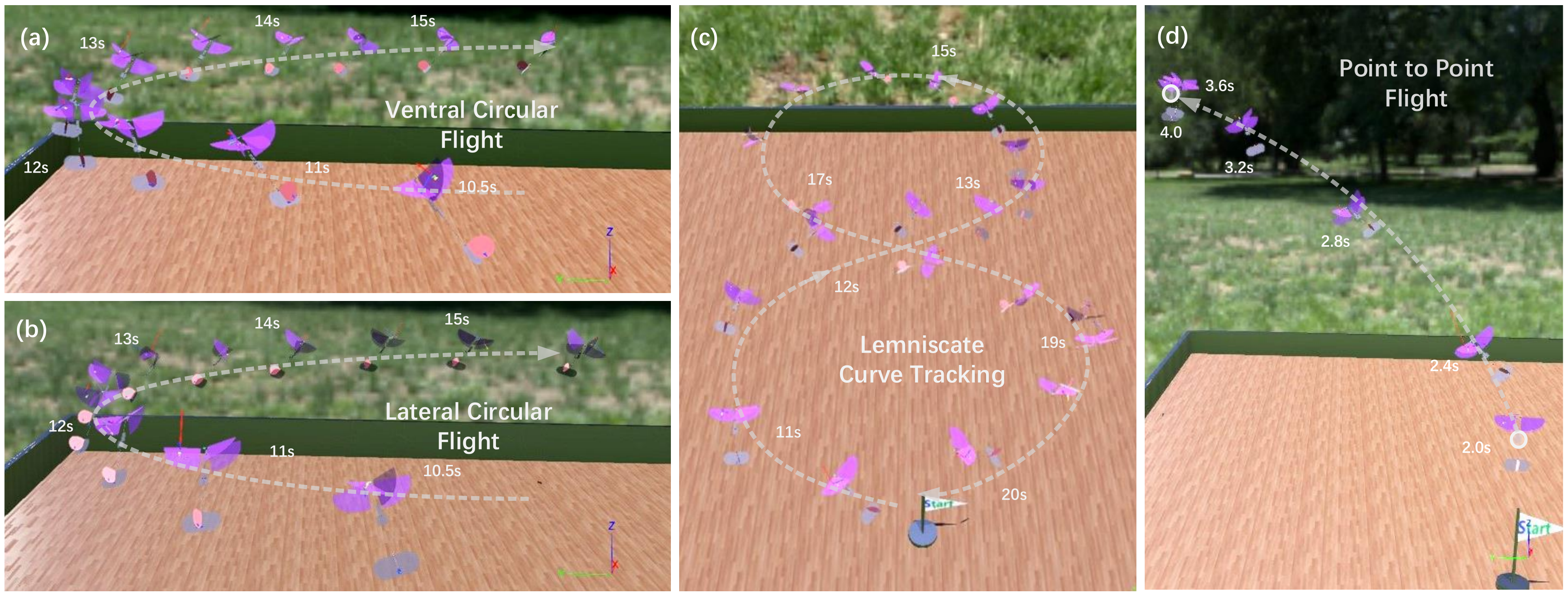}\\
      \caption{Snapshots of different trajectory tracking tasks: (a) ventral circular flight, (b) lateral circular flight,
      (c) Lemniscate curve tracking, (d) point-to-point flight.}
      \label{figure:Cir}
\end{figure}

The simulations are performed with 4 different tasks, namely ventral circular flight (v-Cir.), lateral circular flight (l-Cir.),  Lemniscate curve tracking (Lem.) and point-to-point flight (P2P), where time is measured in seconds:
\begin{align}
{\rm{Cir}}{\rm{.:}}&\left\{ {\begin{array}{l}
{{p_{dx}}\left( t \right) = 2 - 2\cos \left( {0.2\pi t} \right)}\\
{{p_{dy}}\left( t \right) = 2\sin \left( {0.2\pi t} \right)}\\
{{p_{dz}}\left( t \right) = 0}
\end{array}} \right.\nonumber\\
{\rm{Lem}}{\rm{.:}}&\left\{ {\begin{array}{l}
{{p_{dx}}\left( t \right) = 2 - 2\cos \left( {0.2\pi t} \right)/\left( {1 + {{\sin }^2}\left( {0.2\pi t} \right)} \right)}\\
{{p_{dy}}\left( t \right) = 2\sin \left( {0.2\pi t} \right)/\left( {1 + {{\sin }^2}\left( {0.2\pi t} \right)} \right)}\\
{{p_{dz}}\left( t \right) = 0}
\end{array}} \right.\nonumber\\
{\rm{P2P}}:&\left\{ {\begin{array}{l}
{{p_{dx}}\left( t \right) = 0}\\
{{p_{dy}}\left( t \right) = 0}\\
{{p_{dz}}\left( t \right) = 0}
\end{array}} \right.,t \le 2~~~\left\{ {\begin{array}{*{20}{c}}
{{p_{dx}}\left( t \right) = 2}\\
{{p_{dy}}\left( t \right) = 2}\\
{{p_{dz}}\left( t \right) = 2}
\end{array}} \right.,2 < t \le 8\nonumber\\
&\left\{ {\begin{array}{*{20}{c}}
{{p_{dx}}\left( t \right) = 0}\\
{{p_{dy}}\left( t \right) = 0}\\
{{p_{dz}}\left( t \right) = 0} \nonumber
\end{array}} \right., t > 8
\end{align}
where $\bm p_d(t) = \left[ {{p_{dx}}\left( t \right)~{p_{dy}}\left( t \right)~{p_{dz}}\left( t \right)} \right] ^ \top {\rm m}$ 
is fed into the trajectory tracking controller as the desired trajectory. 
In tasks of v-Cir., Lem., and P2P, $\bm d_T$ is set as $[1~0~0]^\top$, while in l-Cir., $\bm d_T$ is set as $[0~1~0]^\top$.
The robot snapshots in stable tracking with respect to these trajectories are shown in Fig. \ref{figure:Cir}.
Parameters for these three controllers are tuned to 
achieve their own optimal performances, and are fixed for all 4 simulation tasks, which are collectively shown in TABLE \ref{tab:TTpara}. 

\begin{figure}[t]
      \centering
      \includegraphics[width=3in]{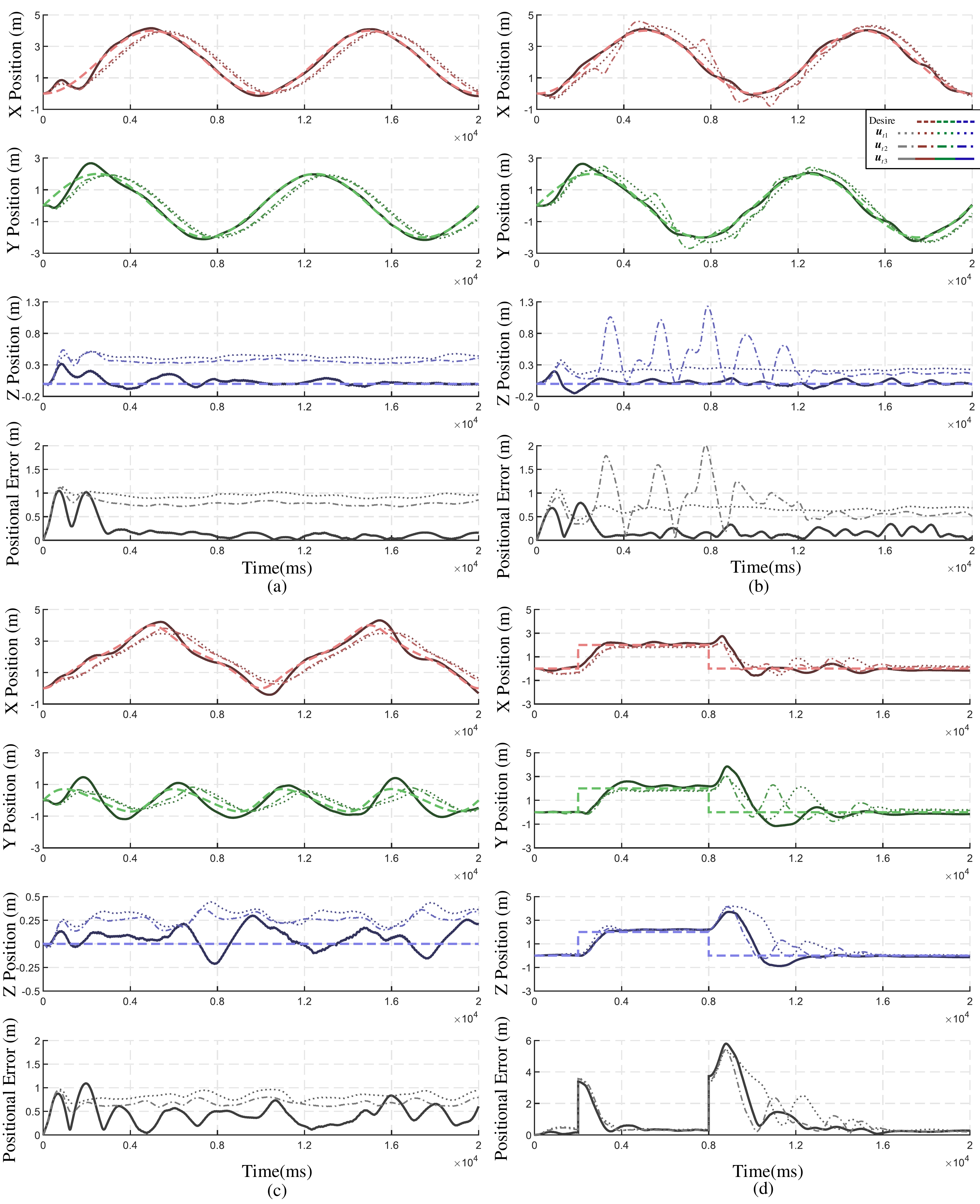}\\
      \caption{Trajectory tracking simulation results with 3 different control laws $\bm u_{t1}$, $\bm u_{t2}$ and $\bm u_{t3}$, within 4 different positional trajectories : (a) ventral circular flight, (b) lateral circular flight,
      (c) Lemniscate curve tracking, (d) point-to-point flight.}
      \label{figure:PosShow}
\end{figure}
Based on the observation of the simulation results shown in Fig. \ref{figure:PosShow}, and further the steady state performance shown in TABLE \ref{tab:TTRes}, we can conclude that the control strategy $\bm u_{t1}$ significantly superiors over the other two comparisons, with at least $25\%$ RMS drop, and approximately $50\%$ RMS drop in average.
Furthermore, comparing the simulation results between (a) and (b), (a) and (c), we can find that the tracking performance deteriorates in lateral flight, or when tracking other large curvature trajectories.
When the desired trajectory is discontinuous, as show in (d), controllers can barely achieve the tracking mission,
which reveals that trajectory planning is needed, however, beyond the scope of this paper.




\section{Experimental Validation}
\begin{figure}[t]
      \centering
      \includegraphics[width=2.8in]{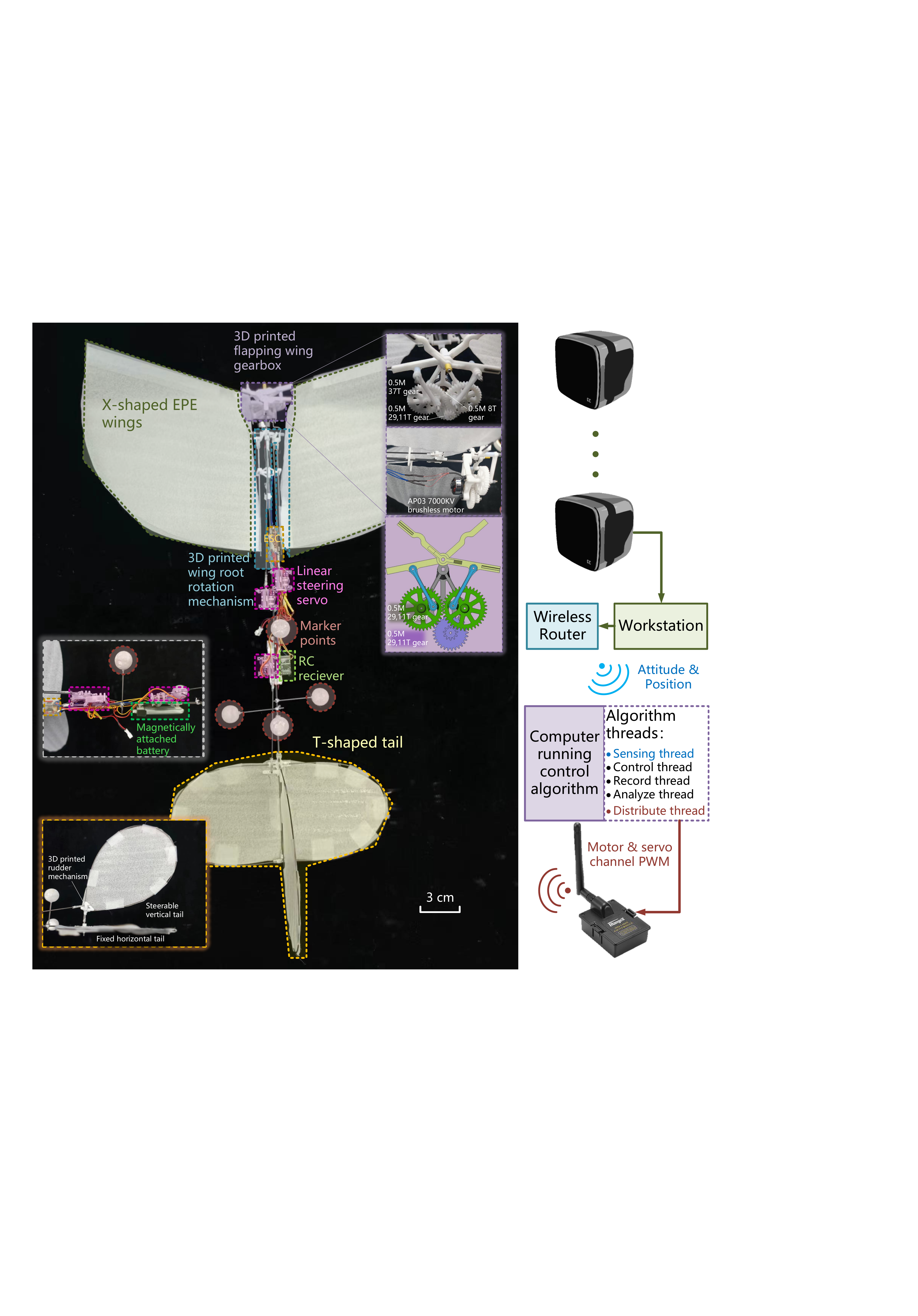}\\
      \caption{Flapping wing robot platform used in real flight experiment.}
      \label{figure:Realflap}
\end{figure}

In order to validate the proposed flapping wing robot simulation platform, we conduct a series of practical real flights.
These experiments are performed under a Qualisys motion capture arena with 48 Arqus A12 cameras online. 
And a self-made 29g flapping wing robot possessing the 
identical configuration as the robot in the above simulations, is tested.
The controller is programmed with Python in a multithread fashion, which endows it with expansibility, 
and can be safely run on a non-real-time system with a desired frequency of 100Hz. 
We attach 4 tracking mark points to the vehicle to estimate its attitude and position. The ``DIY multi-protocol TX
module'' is used to transmit the control signals to the receiver, with a desired frequency of 50Hz. 
Since the system is non-real-time, the computing and transmitting frequencies are not precisely fixed at their desired ones.
To this end, an analyzing thread is performed to analyze other thread frequencies online, and 
further fine-tune the computations involved time.
The overall platform constitution is shown in Fig.~\ref{figure:Realflap}.

\begin{figure}[t]
      \centering
      \includegraphics[width=2.8in]{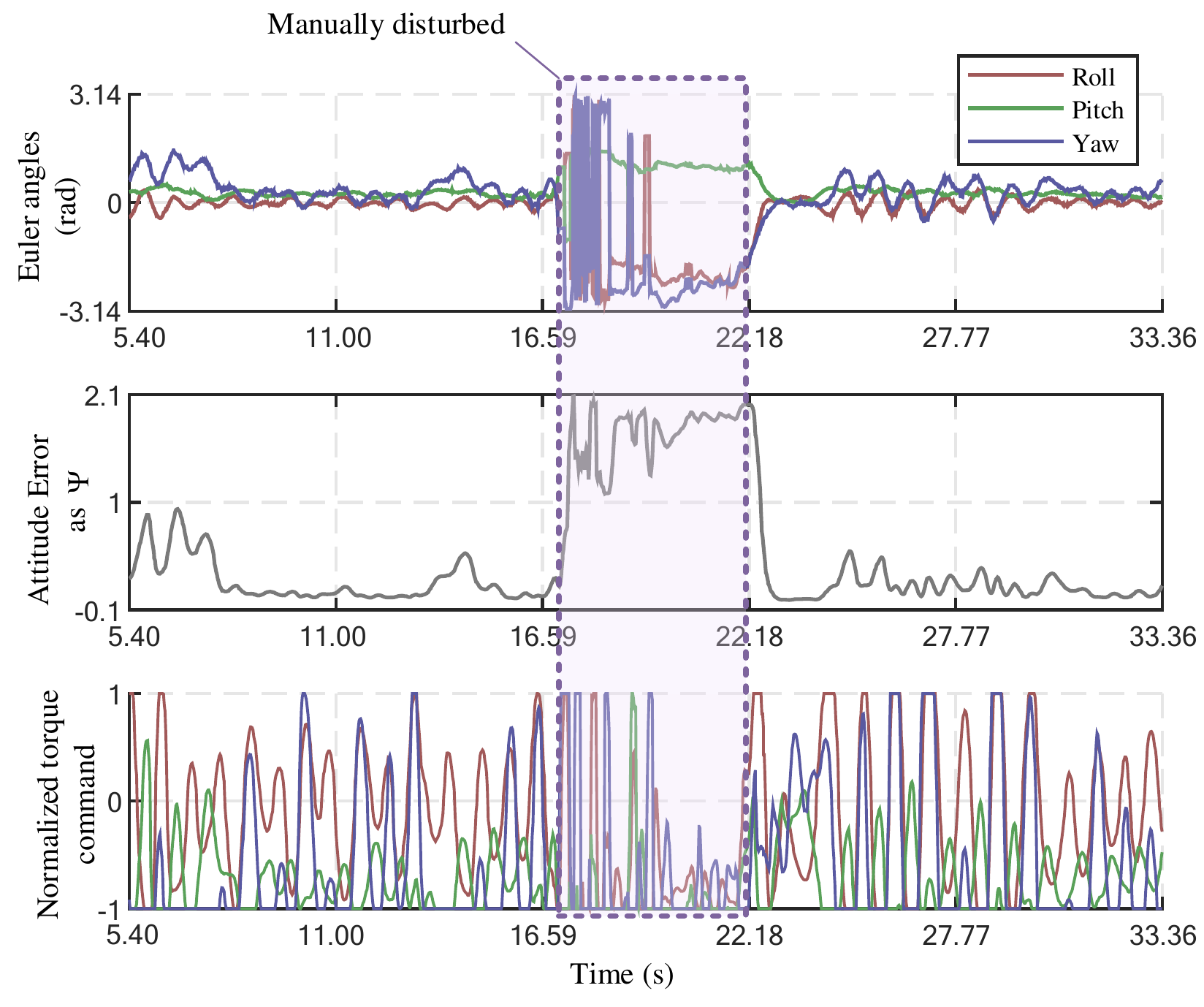}\\
      \caption{Flapping wing robot behavior equipped with control strategy $\bm \tau_3$ when stabilizing the robot to the desired attitude ${\rm{diag}}\left( {\left[ {1~1~1} \right]^\top} \right)$ in real flight experiment.}
      \label{figure:ExpDia}
\end{figure}

The attitude stabilization experiments are performed with three different attitude controllers \eqref{eq:pro_at_con} with the
 desired attitude set as ${\rm{diag}}\left( {\left[ {1~1~1} \right]^\top} \right)$ \footnote{See https://github.com/Chainplain/Flapping\_wing\_TrajControlExperiment.}.
We use exactly the same control laws $\bm \tau_3$ and parameters provided in TABLE \ref{tab:ATRes} except that output channel mappings are modified to adapt the RC receiver signal, 
using pulse width modulation technique (PWM).
Therefore, we can directly check the simulation accuracy by comparing simulation/experiment results.

The experimental result obtained from a flight experiment is shown in Fig.~\ref{figure:ExpDia}.
Based on the result, we can conclude that the proposed control law and parameters tuned in simulations can also 
be directly applied to the real flight with satisfactory performance. 
Specifically, the robot can adjust its attitude to the desired attitude, even after
manually disturbing, which manifests system robustness. 
Furthermore, comparing with the simulation in \ref{figure:AttitudeShow}-(e), we can find that
the stabilizing time length in real flight experiments is similar to that in simulations.

Further comparison experiments are also conducted to corroborate the controller comparison results
based on simulations.  The comparison experiments result are shown in TABLE \ref{tab:Exp}.
And, we find that the proposed controller also has better steady state performance
than the other two, which validates the conclusion made by observing our simulation results.
\begin{table}[t]
\caption{Attitude stabilization experiment  \\ steady state performance measured in $\Psi$}
  \centering
  \label{tab:Exp}
\begin{tabular}{lccc}
\toprule
\multirow{2}{*}{Indicators} & \multicolumn{3}{c}{Control strategies}                                                                            \\ \cline{2-4} 
                            & \multicolumn{1}{l}{$\bm \tau_{1}$} & \multicolumn{1}{l}{$\bm \tau_{2}$} & \multicolumn{1}{l}{$\bm \tau_{3}$ Pro.} \\ \hline
\multicolumn{1}{c}{MAX}     & 0.4359                             & 0.5331                             & \textbf{0.2730}                                 \\
\multicolumn{1}{c}{RMS}     & 0.1250                             & 0.2012                             & \textbf{0.0941}                                 
\\ \bottomrule
\end{tabular}
\end{table}

\section{Conclusion}
In this paper, we propose a novel robotic application-oriented flapping wing simulation platform. 
The platform is used to investigate the dynamic characteristic, such as the passive wing
rotation and the wing-tail interaction phenomena. 
The trade-off between computation simplicity and fidelity is carefully dealt with, such that
most robotic tasks can be simulated, meanwhile, in application-oriented simulations, the 
forces and torques actuated on each blade element can be obtained, which are both rarely achieved in
existing simulation tools. Moreover, the
attitude tracking control and the positional trajectory
tracking tasks as well as their comparison tests are successfully performed on the proposed simulation platform.
Last but not least, the real flight experiments are also completed successfully by directly applying exactly the same algorithms
and parameters used in simulations, which indicates that, to some extent, the gap between flapping wing flight simulation and real robotic application flight is bridged.
In the future, we will fully use this simulation tool to develop flapping wing robots with various scales and configurations. 
Learning based and data driven algorithms will also be applied to explore their extensibility and generalization performance.



%

\appendices
\section{Flapping Wing Statistics of Oscillations}
The imperative need of characterizing the flapping wing oscillation in the sense of dynamics 
impels us to define an intuitive overall statistic.

First of all, periodically computed variance (PV) and standard deviation (PSD) are not suitable, 
because flapping wing forces and torques with large PV and PSD may also induce relatively 
steady body kinematics. 
For example, higher flapping frequency will generate smoother dynamics, however, 
it does not necessarily lead to lower PV or PSD as expected. 

To this end, we propose a novel statistic for flapping wing aerodynamic forces and torques that can capture the oscillating dynamics.
Since the forces and torques are not definitely periodic, consider following the general averaging theory first.

\begin{Theorem}[General averaging \cite{Sanders-2007}]
Consider two dynamics system with the same initial condition:
\begin{equation}
\bm x = \varepsilon {\bm f}\left( {\bm x,t} \right),~~~~\bm x\left( 0 \right) = \bm a
\end{equation}
where $\bm x, \bm a \in D \subset \mathbb R^n $ and $\bm f(\star_1, \star_2) : \mathbb R^n \times \mathbb R \to \mathbb R^n  $.
\begin{equation}~~~~
\bm z = \varepsilon \bar {\bm f}\left( \bm z \right),~~~~\bm z\left( 0 \right) = \bm a~~
\end{equation}
where $\bm z \in D \subset \mathbb R^n $ and $\bm f(\star) : \mathbb R^n  \to \mathbb R^n  $.

Suppose the following two issues are satisfied. 
\begin{enumerate}
\item $\bm f(\star_1, \star_2) $ is a KBM-vector field with average  $\bar {\bm f}(\star)$ and order function $\delta(\varepsilon)$.
\item Trajectory $\bm z (t)$ belongs to an interior set of $D$ on the time scale $1/ \varepsilon$.
\end{enumerate}
Then the two systems satisfy
\begin{equation}
\label{diff}
\bm x(t) - \bm z(t) = \mathcal O (\sqrt {\delta \left( \varepsilon  \right)} )
\end{equation}
as $\varepsilon  \to 0$ on the time scale $1/ \varepsilon$, where 
\begin{equation}
\delta \left( \varepsilon  \right) = \mathop {\sup }\limits_{\bm x \in D} \mathop {\sup }\limits_{t \in \left[ {0,L/\varepsilon } \right)} \varepsilon \left\| {\int_0^t {\left[ { \bm f\left( {\bm x,s} \right) - \bar{\bm f}\left( \bm x \right)} \right]{\rm{d}}s} } \right\|
\end{equation}
\end{Theorem}
Immediately, according to this difference bound between the oscillating system and the averaged system, we can design the statistic describing flapping wing oscillation degree as \eqref{eq:oscillation}.

\end{document}